\documentclass{article}

\usepackage{microtype}
\usepackage{graphicx}
\usepackage{subfigure}
\usepackage{booktabs} 
\usepackage{enumitem}
\usepackage{natbib}
\usepackage{latexsym,amsmath,amssymb,amsthm,eucal,bbm,color}
\usepackage{url}
\usepackage{comment}
\usepackage{float}
\usepackage{tcolorbox}
\usepackage{booktabs}
\usepackage{blindtext}
\usepackage{hyperref}
\hypersetup{
    colorlinks,
    linkcolor={red!50!black},
    citecolor={blue!50!black},
    urlcolor={blue!80!black}
}
\usepackage{multirow}
\usepackage{booktabs}
\usepackage{array}
\usepackage{soul}
 
\usepackage[normalem]{ulem}
\usepackage{stackengine}
\usepackage{parskip}
\usepackage{bm}
\usepackage{balance}
\usepackage{comment}
\usepackage{soul}
\usepackage{listings}
\usepackage[version=4]{mhchem}
\usepackage{empheq}
\usepackage{mdframed}
\usepackage{nomencl,etoolbox,ragged2e,siunitx}
\usepackage{tikz}
\usetikzlibrary{shapes,arrows}
\usetikzlibrary{er,positioning}

\usetikzlibrary{fit,positioning}
\tikzstyle{block} = [rectangle, draw, fill=blue!20, 
    text width=12.8em, text centered, rounded corners, minimum height=4em]
\tikzstyle{line} = [draw, -latex']
\tikzstyle{cloud} = [draw, ellipse,fill=red!20, node distance=3cm,
    minimum height=2em]
\usepackage{mdframed}

\newtheorem{thm}{Theorem}

\newtheorem{prop}{Proposition}
\newtheorem{cor}{Corollary}
\newtheorem{lemma}{Lemma}
\newtheorem{defn}{Definition}

\newtheorem*{remark*}{Remark}

\newmdtheoremenv{mhyp}{Hypothesis} 
\newmdtheoremenv{mthm}{Theorem}
\newmdtheoremenv{mtheorem}{Theorem}
\newmdtheoremenv{mprop}{Proposition}
\newmdtheoremenv{mcor}{Corollary}
\newmdtheoremenv{mlemma}{Lemma}
\newmdtheoremenv{mdefn}{Definition}
\newmdtheoremenv{mmydef}{Definition}
\newmdtheoremenv{mconj}{Conjecture}
\newmdtheoremenv{mex}{Example}
\newmdtheoremenv{mexercise}{Exercise}
\usepackage{wrapfig}
\usepackage{caption}

\usepackage{tikz}
\usetikzlibrary{calc}
\usepackage{pgfplots}
\usepackage{mwe}
\usepackage[]{algorithm2e}
\usepackage{cancel}
\DeclareMathAlphabet\mathbfcal{OMS}{cmsy}{b}{n}

\allowdisplaybreaks

\def \bx{\boldsymbol{x}}

\def \bz{\boldsymbol{z}}

\def \bv{\boldsymbol{v}}
\def \bp{\boldsymbol{p}}

\def \bu{\boldsymbol{u}}
\def \bG{\boldsymbol{G}}

\def \bv{\boldsymbol{v}}
\def \bb{\boldsymbol{b}}

\def \bq{\boldsymbol{q}}

\def \br{\boldsymbol{r}}
\def \bR{\boldsymbol{R}}

\def \bA{\boldsymbol{A}}

\def \bW{\boldsymbol{W}}

\def \gepsilon{\boldsymbol{\epsilon}}

\def\R{{\mathbb R}}

\def\Indic{\mathbbm{1}}

\newcommand{\rank}{\text{rank}}
\newcommand{\adj}{\text{adj}}
\newcommand{\diag}{\text{diag}}

\newcommand{\qq}{\vspace*{-2mm}}

\usepackage{xcolor}

\usepackage{tikz}
\usetikzlibrary{shapes,arrows}
\usetikzlibrary{er,positioning}

\usetikzlibrary{fit,positioning}
\tikzstyle{block} = [rectangle, draw, fill=blue!20, 
    text width=12.8em, text centered, rounded corners, minimum height=4em]
\tikzstyle{line} = [draw, -latex']
\tikzstyle{cloud} = [draw, ellipse,fill=red!20, node distance=3cm,
    minimum height=2em]
    
\usepackage{enumitem}   
\usepackage{bm}
\usepackage[nomessages]{fp}
\usepackage{hyperref}

\usepackage{float}
\usepackage[accepted]{icml2020}

\icmltitlerunning{Max-Affine Spline Insights into Generative Deep Networks}

\begin{document}

\twocolumn[
\icmltitle{Max-Affine Spline Insights into Deep Generative Networks}

\begin{icmlauthorlist}
\icmlauthor{Randall Balestriero}{to}
\icmlauthor{Sebastien Paris}{goo}
\icmlauthor{Richard G. Baraniuk}{to}
\end{icmlauthorlist}

\icmlaffiliation{to}{ECE Department, Rice University, TX, USA}
\icmlaffiliation{goo}{LSIS, UTLN, La Garde, France}

\icmlcorrespondingauthor{Randall Balestriero}{randallbalestriero@gmail.com}
\icmlkeywords{Generative Deep Networks, Machine Learning, Manifold, Dropout, Multimodal Distributions}

\vskip 0.3in
]



\printAffiliationsAndNotice{}  

\begin{abstract}
We connect a large class of Generative Deep Networks (GDNs) with spline operators in order to derive their properties, limitations, and new opportunities.
By characterizing the latent space partition, dimension and angularity of the generated manifold, we relate the manifold dimension and approximation error to the sample size. 
The manifold-per-region affine subspace defines a local coordinate basis; we provide necessary and sufficient conditions relating those basis vectors with disentanglement. 
We also derive the output probability density mapped onto the generated manifold in terms of the latent space density, which enables the computation of key statistics such as its Shannon entropy. 
This finding also enables the computation of the GDN likelihood, which provides a new mechanism for model comparison as well as providing a quality measure for (generated) samples under the learned distribution. 
We demonstrate how low entropy and/or multimodal distributions are not naturally modeled by DGNs and are a cause of training instabilities.
\end{abstract}

\section{Introduction}

Deep Generative Networks (DGNs), which map a low-dimensional latent variable $\bz$ to a higher-dimensional generated sample $\bx$, have made enormous leaps in capabilities in recent years. 
Popular DGNs include Generative Adversarial Networks (GANs) \cite{goodfellow2014generative} and their variants  \cite{dziugaite2015training,zhao2016energy,durugkar2016generative,arjovsky2017wasserstein,mao2017least,yang2019diversitysensitive};
Variational Autoencoders 
\cite{kingma2013auto} and their variants   \cite{fabius2014variational,van2017neural,higgins2017beta,tomczak2017vae,davidson2018hyperspherical}; and 
flow based models such as NICE \cite{dinh2014nice}, 
Normalizing Flow \cite{rezende2015variational}, and their variants \cite{dinh2016density,grathwohl2018ffjord,kingma2018glow}.

While DGNs are easy to describe and analyze locally in terms of simple affine operators and scalar nonlinearities, a general framework for their {\em global structure} has remained elusive. 
In this paper, we take a step in the direction of a better theoretical understanding of DGNs constructed using continuous, piecewise affine nonlinearities by leveraging recent progress on {\em max-affine spline operators} (MASOs) \cite{balestriero2018spline}. 
Our main contributions are as follows;

{\bf [C1]}
We characterize the {\em piecewise-affine manifold structure} of the generated samples, including its {\em intrinsic dimension} (Section~\ref{sec:manifold_dimension}), which sheds new light on the impact of techniques like dropout (Section~\ref{sec:dropout}) and provides practionioners with sensible design principles for constructing a desired manifold.

{\bf [C2]}
We characterize the {\em local coordinates of the generated manifold} and the inverse mapping from data points $\bx$ back to latent variables $\bz$ (Section~\ref{sec:local_coordinates}), which provides new necessary and sufficient conditions for {\em disentanglement}, {\em interpretability} and new links between DGNs and {\em adaptive basis methods} (Section~\ref{sec:adaptive_basis}).  
By characterizing the angles between adjacent local affine regions, we demonstrate how weight sharing in a DGN heavily constrains the {\em curvature} of the generated manifold despite the fact that the DGN might be tremendously overparameterized (Section~\ref{sec:curvature}).

{\bf [C3]}
We provide a DGN input-output formula that enables us to derive the analytical {\em probability density on the generated manifold} that is induced by the latent space  (Section~\ref{sec:output_density}).
We use this result to derive {\em Normalizing Flows} (NMFs) form first principles and highlight how the DGN design, $\bx \mapsto \bz$  (most NMFs) versus $\bz \mapsto \bx$ (DGNs) allow for either fast likelihood computation and slow sampling or vice-versa (Section~\ref{sec:nmf}). Finally, the Shannon entropy of the output density provide a new lens through which to study the difficulty of generating multidimensional, low-entropy distributions using DGNs (Section~\ref{sec:difficulty}).

Reproducible code for the various experiments and figures is be provided on Github\footnote{\url{https://github.com/RandallBalestriero/GAN.git}}.

\section{Background}
\label{sec:background}

{\bf Deep Networks.}
A deep (neural) network (DN) is an operator $f_\Theta$ with parameters $\Theta$ that maps the input $\bz\in\R^S$ to the output $\bx\in \R^D$ by composing $L$ intermediate {\em layer} mappings $f_{\ell}$, $\ell=1,\dots,L$, that combine affine and simple nonlinear operators such as the {\em fully connected operator} (simply the affine transformation defined by the weight matrix $\bW_{\ell}$ and bias vector $\bb_{\ell}$), {\em convolution operator} (with circulent $\bW_{\ell}$), {\em activation operator} (applying a scalar nonlinearity such as the ubiquitous ReLU), or {\em pooling operator}. 
Precise definitions of these operators can be found in \cite{goodfellow2016deep}. 
We will omit $\Theta$ for conciseness unless it is needed.

We precisely define a layer $f_{\ell}$ as comprising a single nonlinear operator composed with any (if any) preceding linear operators that lie between it and the preceding nonlinear operator.
Each layer $f_{\ell}$ transforms its input {\em feature map}  $\bv_{\ell-1}\in\R^{D_{\ell-1}}$
into an output feature map $\bv_{\ell}\in\R^{D_{\ell}}$
with the initializations $\bv_{0}:=\bz$,
$D_{0}=S$, and $\bv_{L}:=\bx, D_{L}=D$.
In this paper, we focus on DGNs, where $S<D$, 
$\bz$ is interpreted as a latent representation, and $\bx$ is the generated data, e.g, a time-serie or image.
The feature maps $\bv_{\ell}$ can be viewed equivalently as signals, flattened column vectors, or tensors depending on the context.

{\bf Max-Affine Spline Deep Networks.}
A $K$-dimensional {\em max-affine spline operator} (MASO) concatenates $K$ independent {\em max-affine spline} (MAS) functions, with each MAS formed from the point-wise maximum of $R$ affine mappings
\cite{magnani2009convex,hannah2013multivariate}. 
Given an input vector $\bu$, the output of a MASO is given by
\begin{align}
{\rm MASO}(\bu;\{\bA_r,\bb_r\}_{r=1}^R) = \max_{r=1,\dots,R} \bA_{r}\bu + \bb_{r},
\end{align}
where $\bA_r \in \mathbb{R}^{D_{\ell}\times D_{\ell-1}}$ are the slopes and 
$\bb_r \in \mathbb{R}^{D_{\ell}}$ 
are the offset/bias parameters 
$\forall r$ and the maximum is taken coordinate-wise.
Note that a MASO is a {\em continuous piecewise-affine} (CPA) operator \cite{wang2005generalization}.

The key background result for this paper is that {\em the layers of DNs (DGNs) constructed from piecewise affine operators (e.g., convolution, ReLU, and max-pooling) are MASOs}
\citet{balestriero2018spline,reportRB}; hence a DN (DGN) is a composition of MASOs.
For example, a layer comprising a fully connected operator with weights $\bW_{\ell}$ and biases $\bb_{\ell}$ followed by a ReLU activation operator has parameters $R=2, \bA_1=\bW_{\ell},\bA_2=\mathbf{0}, \bb_{1}=\bb_{\ell}, \bb_{2}=\mathbf{0}$.

The piecewise-affine spline interpretation provides a powerful global geometric interpretation of a DN (DGN) in which it partitions its input space $\R^L$ into polyhedral regions (the set $\Omega$) and then assigns a different, fixed affine transformation to each region.
The partition regions are built up over the layers via a {\em subdivision} process and are closely related to Voronoi and power diagrams \cite{balestriero2019geometry}.

\section{The Generated Manifold of a DGN}
\label{sec:manifold_dimension}

In this section we study the properties of the mapping 
$\bG_{\Theta}:\mathbb{R}^{S}\rightarrow \mathbb{R}^{D}$
of a deep generative network (DGN) comprising $L$ piecewise-affine MASO layers. 

\subsection{Input Space Partition and Region Mapping}
\label{sec:maso_gdn}

While our approach holds for arbitrary piecewise affine layers, for concreteness of exposition, we will focus on nonlinearities with $R=2$ (e.g., ReLU, leaky ReLU, absolute value). 
In all such cases, the state of the nonlinearity can be encoded as a value from $\{\alpha, 1\}$ with $\alpha=0$ for ReLU, $\alpha=-1$ for absolute value and $\alpha>0$ for leaky-ReLU. At layer $\ell$, observing an input $\bv_{\ell-1}$ defines the state of the layer nonlinearities and in turn defines the ``piece'' of the layer MASO used to produce the output $\bv_{\ell}$. 
We call the nonlinearity's state its {\em code} $\bq_{\ell}(\bv_{\ell-1})\in \{\alpha, 1\}^{D_{\ell}}, \bv_{\ell-1} \in \mathbb{R}^{D_{\ell-1}}$ with
\begin{align}
    [\bq_{\ell}(\bv_{\ell-1})]_i = \begin{cases}
    \alpha, & [\bW_{\ell}\bv_{\ell-1}+\bb_{\ell}]_i \leq 0\\
    1, & [\bW_{\ell}\bv_{\ell-1}+\bb_{\ell}]_i > 0
    \end{cases}
\end{align}
leading to the simple forward formula for the layer
\begin{align}
\label{eq:layer_mapping}
   f_{\ell}(\bv_{\ell-1})=\diag(\bq_{\ell}(\bv))(\bW_{\ell}\bv+\bb_{\ell}).
\end{align}
We concatenate the per-layer codes into $\bq(\bz)=[\bq_{1}(\bz)^T,\dots,\bq_{L}(\bz)^T]^T \in \{\alpha, 1\}^{\prod_{l=1}^{L}D_l}$ with $\bz \in \mathbb{R}^S$ the DGN input and $\bq_{\ell}(\bv_{\ell-1})$ abbreviated as $\bq_{\ell}(\bz)$.

\begin{defn}
\label{def:partition}
A {\bf partition region} $\omega_{\mathbf{k}}$ of the DGN input space partition $\Omega$ is defined as the input space region for which the MASO states $\bq(\cdot)$ are identical
\begin{align}
    \omega_{\mathbf{k}} =& \left\{\bz \in \mathbb{R}^{S}: \bq(\bz)={\mathbf{k}}\right\}, \forall \mathbf{k}\in \{\alpha, 1\}^{\prod_{\ell=1}^{L}D_{\ell}},\label{eq:region}\\
    \Omega = & \left\{\omega_{\mathbf{k}}, \mathbf{k}\in \{\alpha, 1\}^{\prod_{\ell=1}^{L}D_{\ell}}\right\}\setminus \emptyset. \label{eq:partition}
\end{align}
\end{defn}
\qq

Note that $\cup_{\omega \in \Omega}\;  \omega=\mathbb{R}^S$ and $\forall (\omega,\omega') \in \Omega^2, \omega \not = \omega', \omega^{\circ} \cap \omega^{'\circ} = \emptyset$, with $(\cdot)^{\circ}$ the interior operator \cite{munkres2014topology}. 

Since $\bG$ is formed from a composition of MASOs, it is itself a CPA with a fixed affine mapping over each region $\omega \in \Omega$.

As a result, the generator $\bG$ maps each $S$-dimensional convex latent space region $\omega \in \Omega$ to the convex affine subspace $\bG(\omega) \subset \mathbb{R}^D$ as
\begin{align}
    \label{eq:region_mapping}
    \forall \omega \in \Omega, \;\;\bG(\omega) = \{\bA_{\omega} \bz+ \bb_{\omega}, \bz \in \omega \} \subseteq \mathbb{R}^{D},
\end{align}
with $\bA_{\omega}$ and $\bb_{\omega}$ obtained by composing (\ref{eq:layer_mapping}) and distributing the terms; we will call this the {\em generated manifold}.

\begin{prop}
\label{prop:input_region_convexity}
A DGN $\bG$ comprised of MASO layers is a CPA operator with input space partition $\Omega$ given by (\ref{eq:partition}).
Each region $\omega\in\Omega$ is a convex polytope in $\mathbb{R}^S$ that is affinely mapped to the affine subspace $\bG(\omega)$ (recall (\ref{eq:region_mapping}), a convex polytope in $\mathbb{R}^D$ given by (\ref{eq:region_mapping}).
(Proof in Appendix \ref{proof:prop:input_region_convexity}.)
\end{prop}
\qq

Analytically, the form of (\ref{eq:region_mapping}) can be obtained directly by composing the per layer mappings from (\ref{eq:layer_mapping}), distributing and rearranging the terms into the slope and bias terms of the affine mapping.
Computationally, the affine parameters $\bA_{\omega}, \bb_{\omega}$ can be obtained efficiently in one of the two following ways. On the one hand, if one possesses an input $\bz$ belonging to the desired region $\omega$, then we simply perform
\begin{align}
    \bA_{\omega}  = \nabla _{\bz} \bG(\bz),\;\; \bb_{\omega} = \bG(\bz) - \bA_{\omega}\bz.\label{eq:compute_Ab}
\end{align}
On the other hand, in the case where one has access to the code $\bq(\omega)$ of the region (as opposed to a point in the region), one can directly impose the nonlinearity states (defined by $\bq(\omega)$) on the DGN mapping. Once the nonlinearities are fixed, one can feed an arbitrary input $\bz \in \mathbb{R}^S$ and compute the affine parameters as in (\ref{eq:compute_Ab}) on the fixed DGN.

We can extend (\ref{eq:region_mapping}) to the entire domain of the generator via 
\begin{align}
    \bG(\text{supp}(\bp_{\bz})) =\hspace{-0.1cm} \bigcup_{\omega \in \Omega} \bG(\omega \cap \text{supp}(\bp_{\bz})) ,\label{eq:generator_mapping}
\end{align}
with $\bp_{\bz}$ the probability distribution on the latent space that generates $\bz$, and where $\bG(\text{supp}(p_{\bz}))$ denotes the image of $\bG$ by considering all inputs $\bz \in \text{supp}( p_{\bz})$ with nonzero probability, e.g., $\mathbb{R}^S$ if the latent distribution is a Gaussian, and the $S$-dimensional hypercube if it is a standard Uniform distribution. 
Thus, the generated manifold (\ref{eq:generator_mapping}) combines the per-region affine transformations of the input space partition per (\ref{eq:region_mapping}).
With this formulation, we now characterize the intrinsic dimension of the per-region and overall manifold mapping of $\bG$.

\subsection{Generated Manifold Intrinsic Dimension}

We now turn into the intrinsic dimension of the per-region affine subspaces $\bG(\omega))$ that comprise the generated manifold. In fact, as per (\ref{eq:generator_mapping}), its dimension depends not only on the latent dimension $S$ but also on the per layer parameters.

\begin{lemma}
\label{lemma:upper_bound}
The intrinsic dimension of the affine subspace $\bG(\omega))$ (recall (\ref{eq:generator_mapping})) has the following upper bound
\begin{align*}
    \dim(\bG(\omega)) \leq \min \Big(S, \min_{\ell} \big( {\rm rank}\left( {\rm diag}(\bq_{\ell}(\omega))\bW_{\ell}\right)\big) \Big).
\end{align*} 
(Proof in Appendix \ref{proof:upper_bound}.)
\end{lemma}
\qq

We make three observations.  
First, we see that the choice of the nonlinearity (i.e., the choice of $\alpha$) and/or the choice of the per-layer dimensions (i.e., the ``width'' of the DGN) are the key elements controlling the upper bound of $\dim(\bG)$. 
For example, in the case of ReLU ($\alpha=0$) then $\dim(\bG(\omega))$ is directly impacted by the number of $0$s in the codes $\bq_{\ell}(\omega)$ of each layer in addition of the rank of $\bW_{\ell}$; this sensitivity does not occur when using other nonlinearities ($\alpha \not = 0$). Second, ``bottleneck layers'' impact directly the dimension of the subspace and thus should also be carefully picked based on the a priori knowledge of the target manifold intrinsic dimension. 
Third, we obtain the following condition relating the per-region dimension to the bijectivity and surjectivity of the mapping. The latter should be avoided in DGNs, since it implies that multiple different latent vectors will generate the same output sample.

\begin{prop}
\label{prop:bijective}
A DGN is bijective on $\omega$ iff $\dim(\bG(\omega)) = S,\forall \omega \in \Omega$ and surjective iff $\exists \omega \in \Omega \text{ s.t. }\dim(\bG(\omega))<S$.
A DGN is bijective on $\text{supp}(\bp_{\bz})$ iff it is bijective for each region $\omega \in \Omega$ and $\bG(\omega)\cap \bG(\omega') = \emptyset, \forall \omega \not = \omega'$. (Proof in Appendix \ref{proof:bijective}.)
\end{prop}
\qq

\subsection{Application: Effect of Dropout/Dropconnect}
\label{sec:dropout}

\begin{figure}[t]
    \centering
    \begin{minipage}{0.45\linewidth}
    \includegraphics[width=1\linewidth]{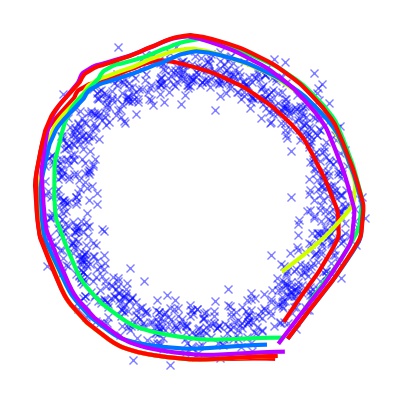}
    \end{minipage}
    \begin{minipage}{0.54\linewidth}
    \caption{\small Demonstration of a GAN DGN trained on a circle dataset. Each line is the learned piecewise linear manifold generated by the DGN; each color corresponds to a different realization of dropout noise. Dropout turns a DGN into a finite ensemble of DGNs with the same or lower intrinsic dimension.
    \normalsize}
    \label{fig:dropout_2d}
    \end{minipage}
\end{figure}

Noise techniques, such as dropout \cite{wager2013dropout} and dropconnect \cite{wan2013regularization}, alter the per-region affine mapping in a very particular way that we now characterize. 

Dropout and dropconnect techniques apply a multiplicative binary noise onto the feature maps and/or the weights; the multiplicative noise $\epsilon_{\ell} \in \{0,1\}^{D_{\ell}}$ is typically an iid Bernoulli random variable.  \cite{isola2017image}. 
To characterize how this noise impacts the DGN mapping, denote the generator $\bG$ equipped with random dropout/dropconnect by $\widetilde{\bG}$, and the generator in which the noise realization is fixed by $\bG(\bz|\gepsilon) = \bA_{\omega}(\gepsilon)\bz+\bb_{\omega}(\gepsilon)$ where $\gepsilon$ concatenates the random variables of each layer. Given the mapping form (recall (\ref{eq:layer_mapping})) with per layer parameters $\bW_{\ell},\bb_{\ell}$,
the presence of dropout noise leads to the following noisy input-output mapping on a region $\omega$ (recall (\ref{eq:region_mapping})) as
\begin{multline}
    \bG(\bz|\gepsilon) = \left(\prod_{\ell=L}^1\diag(\bq_{\ell}(\omega)\odot \epsilon_{\ell})\bW_{\ell}\right)\bz \\
    + \sum_{\ell=1}^L\left(\prod_{\ell'=L}^{\ell+1}\diag(\bq_{\ell'}(\omega)\odot \epsilon_{\ell'})\bW_{\ell'}\right)\bb_{\ell},\label{eq:dropout}
\end{multline}
where $\odot$ is the Hadamard product and $\bz \in \omega$. (See Appendix \ref{sec:dropconnect} for the dropconnect formula.) 
Thus, the noisy generator actually combines all the above mappings for each noise realisation via
\begin{align}
    \widetilde{\bG}(\text{supp}(p_{\bz})) = \bigcup_{\gepsilon \in \text{supp}(p_{\gepsilon})} \bG(\text{supp}(p_{\bz})|\gepsilon).
    \label{eq:dropout_mapping}
\end{align}

\begin{figure*}[t]
    \centering
    \begin{minipage}{0.05\linewidth}
    \rotatebox{90}{\hspace{0.3cm}$\dim(\bG(\omega))$}
    \end{minipage}
    \begin{minipage}{0.94\linewidth}
    \begin{minipage}{0.24\linewidth}
    \centering
    Dropout 0.1
    \end{minipage}
    \begin{minipage}{0.24\linewidth}
    \centering
    Dropout 0.3
    \end{minipage}
    \begin{minipage}{0.24\linewidth}
    \centering
    Dropconnect 0.1
    \end{minipage}
    \begin{minipage}{0.24\linewidth}
    \centering
    Dropconnect 0.3
    \end{minipage}\\
    \includegraphics[width=\linewidth]{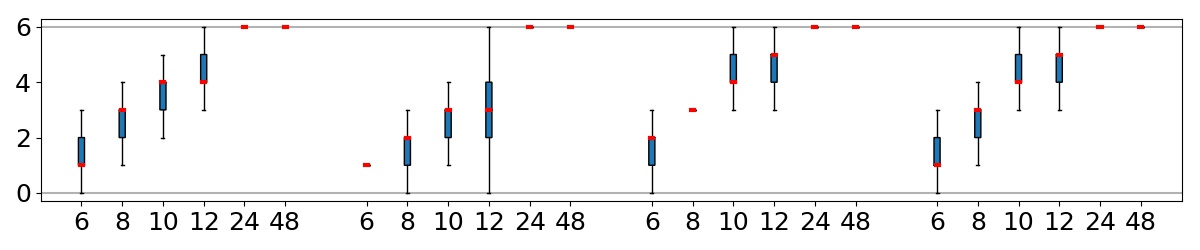}
    \end{minipage}
    \caption{\small{Impact of dropout and dropconnect on the dimension of the noisy generator affine subspaces $\bG(\omega|\epsilon), \forall \omega$ (recall (\ref{eq:dropout_mapping})). We depict two ``drop'' probabilities $0.1$ and $0.3$ for a generator $\bG$ with $S=6$, $D=10$, $L=3$ and varying width $D_1=D_2$ ranging in $\{ 6, 8, 10, 12, 24, 48\}$ (x-axis); note that the architecture limits the dimension to $S=6$. The boxplot represents the distribution of the per-region affine subspace dimensions for $2000$ sampled regions over $2000$ different noise realizations $\epsilon$. 
    We make two observations.
    First, dropconnect tends to preserve the latent dimension $S$ even when the width $D_1,D_2$ is close to $S$. 
    Second, the dropout-induced collection of generators tends to have degenerate dimension (much smaller than $S$) until the width is twice the latent space dimension $(D_\ell \geq 2S)$. As a result, while dropout turns a generator into a collection of generators, those generators will have degenerate dimension unless $\bG$ is much wider that $S$.}}
    \label{fig:dropout}
\end{figure*}

\begin{prop}
\label{prop:mixture}
Multiplicative binary dropout/dropconnect transforms the generator $\bG$ into the union of generators given by (\ref{eq:dropout_mapping}), each with per-region dimension between $0$ and $\max_{\omega}\text{dim}(\bG(\omega))$. (Proof in Appendix \ref{proof:dim_dropout_range}.)
\end{prop}
\qq

From the above, we see that the multiplicative binary noise does not make the generator $\widetilde{\bG}$ dense in its output space, but rather turns $\bG$ into a collection of piecewise linear generators, each corresponding to a different realization of $\gepsilon$ as depicted in Fig.~\ref{fig:dropout_2d}.
Furthermore, each noise realization produces a generator with a possibly different per-region dimension being upper bounded by the dimension of the original generator $\bG$. Also, each induced generator has a possibly different input space partition based on the noise realisation.
On the one hand, this highlights a potential limitation of those techniques for narrow models ($D_{\ell} \approx S$) for which the noisy generators will tend to be degenerate (per-region dimension smaller than $S$), implying surjectivity (recall Prop.~\ref{prop:bijective}). 
On the other hand, when used with wide DGNs ($D_{\ell} \gg S$) much more noisy generators will maintain the same affine subspace dimension that the original generator. The latter is crucial when $S$ is picked a priori to match exactly the true intrinsic dimension. We illustrate the above in Fig.~\ref{fig:dropout}.

\subsection{Application: Optimal Dimension and Training Error Increase}

We now emphasize how the DGN dimension $\tilde{S}\triangleq \max_{\omega}\dim(\bG(\omega))$ impacts the training error loss and training sample recovery. 
We answer the following question: {\em Can a generator generate $N$ training samples from a continuous $S^*$ dimensional manifold if $\tilde{S}<S^*$?} 
Denote the empirical error measuring the ability to generate the data samples by $E^{*}_{N}=\min_{\Theta}\frac{1}{N}\sum_{n=1}^N \min_{\bz }\|\bG_{\Theta}(\bz)-\bx_n \|$.
We now demonstrate and empirically validate that if $\tilde{S}<S^*$ then $E^{*}_N$ increases with $N$ for any data manifold.

\begin{thm}
\label{thm:error_increase}
Given the true intrinsic dimension of an arbitrary manifold $S^*$, any finite DGN with $\tilde{S}<S^*$ will have increasing error $E^*$ with the dataset size $N$ as in $\exists \ell \in \{0, \dots, L\}: D_{\ell}<S^* \implies \forall N>0, \exists N' > N : E^{*}_{N'}>E^*_{N}$. (Proof in Appendix \ref{proof:error_increase}.)
\end{thm}
\qq

In general, it is clear that, for smooth manifolds, $E^*$ increases with $N$ since the DGN is piecewise linear. However, the above result extends this to any manifold, even when the data manifold is as simple as a linear manifold (translated subspace).
We empirically validate this phenomenon in  Fig.~\ref{fig:error_with_n} for the simple case of a linear manifold.
(The experimental details are given in  Appendix~\ref{appendix:error_with_n}.)

\begin{figure}[t]
    \centering
    \begin{minipage}{0.05\linewidth}
    \rotatebox{90}{{\small Minimum error $E^*$}}
    \end{minipage}
    \begin{minipage}{0.93\linewidth}
    \includegraphics[width=1\linewidth]{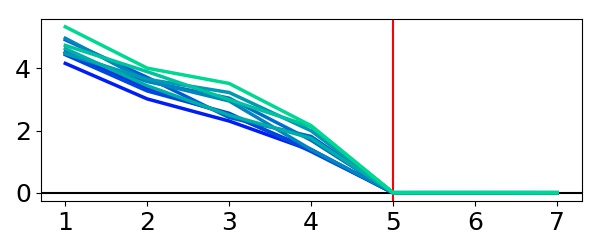}
    \end{minipage}
    \small{DGN latent dimension $S$,\;\; ($S^*$ in red)}
    \caption{Error $E^*$ (y-axis) for a linear manifold with $S^*=5$, for increasing dataset size $N\in \{100, 120, 140, 160, 180, 200, 300, 400, 500, 1000\}$ (blue to green)  for different latent space dimension $S \in \{ 1, 2, 3 ,4 ,5 ,6, 7\}$ (x-axis) which forces $\tilde{S}<S$, the $E^*=0$ line is depicted in black. This demonstrates, as per Thm.~\ref{thm:error_increase}, that whenever $\tilde{S}<S^*$, the training error $E^*$ increases with the dataset size $N$ and is $0$ otherwise whenever $\tilde{S}\geq S^*$.
    }
    \label{fig:error_with_n}
\end{figure}

It is clear that a direct implication of Thm.~\ref{thm:error_increase} is that there does not exist a finite architecture  with $D_{\ell}<D$ for some $\ell$ and parameter $\Theta$ such that a MASO DGN would be bijective with $\mathbb{R}^D$.
The above results are key to understand the challenges and importance of the design of DGN starting with the width of the hidden layers and latent dimensions in conjunction with the choice of nonlinearities and constraints on $\bW_{\ell}$ of all layers.

\section{Manifold Local Coordinates and Curvature}
\label{sec:curvature_coordinate}

We now turn to the study of the local coordinates of the affine mappings comprising a DGN's generated manifold. We then study the coupling between the affine mappings of adjacent regionsto characterize the curvature/angularity of the generated manifold. 

\subsection{Local Coordinate Systems and Inverse Mapping}
\label{sec:local_coordinates}

Recall from (\ref{eq:generator_mapping}) that a DGN is a CPA operator. Inside region $\omega \in \Omega$, points are mapped to the output space affine subspace which is itself governed by a coordinate system or basis.
For the remaining of the section we assume that $\dim(\bG(\omega))=S, \forall \omega \in \Omega$ and thus the columns of $\bA_{\omega}$ are linearly independent. For cases where $\dim(\bG(\omega))<S$ then the following analysis also applies by considering a lower dimensional latent space $S'<S$ and the corresponding sub-network that only depend on the kept latent space dimensions.

\begin{lemma}
\label{lemma:basis}
A basis for the affine subspace $\bG(\omega)$ (recall (\ref{eq:region_mapping})) is given by the columns of $\bA_{\omega}$.
\end{lemma}
\qq

In other words, the columns of $\bA_{\omega}$ form the local coordinate space, and each latent space dimension moves a point in this region by adding to it the corresponding slope column. Prior leveraging this result for latent space characterization, we derive an inverse of the generator $\bG$ that maps any point from the generated manifold to the latent space. This inverse is well-defined as long as the generator is injective, preventing that $\exists \bz_1 \not = \bz_2$ s.t. $ \bG(\bz_1)=\bG(\bz_2)$. Assuming injectivity, the inverse of $\bG$ on a region $\bG(\omega)$ in the output space is obtained by
\begin{align}
    \bG_{\omega}^{-1}(\bx) =\left( \bA^T_{\omega}\bA_{\omega}\right)^{-1}\bA_{\omega}^T(\bx -\bb_{\omega}), \forall \bx \in \bG(\omega),\label{eq:region_inverse}
\end{align}
leading to $\bG_{\omega}^{-1}(\bG(\omega))=\omega, \forall \omega \in \Omega$. Note that the inverse $\left( \bA^T_{\omega}\bA_{\omega}\right)^{-1}$ is well defined as $\bA_{\omega}$ is full column rank since we only consider a generator with $\tilde{S}=S$. 
We can then simply combine the region-conditioned inverses to obtain the overall generator inverse.

\begin{lemma}
\label{lemma:inverse}
The inverse mapping of an injective DGN is the CPA operator mapping 
$\bG(\text{supp}(\bp_{\bz})) \mapsto \text{supp}(\bp_{\bz})$ given by
$
    \bG^{-1}(\bx) =\sum_{\omega \in \Omega}\bG_{\omega}^{-1}(\bx)\Indic_{\{\bx \in \bG(\omega)\}}.
$
(Proof in Appendix \ref{proof:inverse}.)
\end{lemma}
\qq


\subsection{Application: Adaptive Basis and Disentenglement}
\label{sec:adaptive_basis}

As mentioned in the above section, $\bA_{\omega}$ forms a basis of the affine subspace $\bG(\omega)$. The latent vector $\bv$ combines them to obtain the subspace which is then shifted by the bias $\bb_{\omega}$. 
This process is performed locally for each region $\omega$, in a manner similar to an ``adaptive basis'' \cite{donoho1994ideal}.

\begin{figure}[t]
\begin{minipage}{0.05\linewidth}
\rotatebox{90}{\small learned \hspace{0.6cm} initial}
\end{minipage}
\begin{minipage}{0.21\linewidth}
\centering
\tiny
FC GAN\normalsize\\
\includegraphics[width=\linewidth]{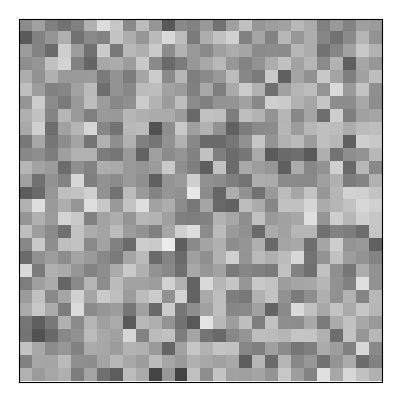}\\
\includegraphics[width=\linewidth]{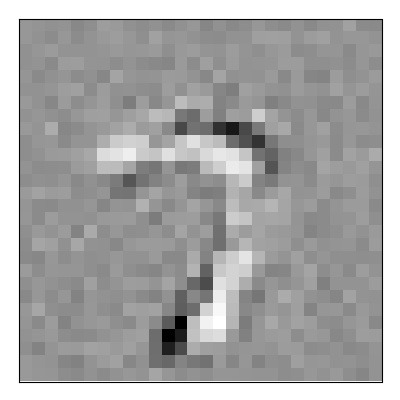}\\
\end{minipage}
\begin{minipage}{0.21\linewidth}
\centering
\tiny
CONV GAN\normalsize\\
\includegraphics[width=\linewidth]{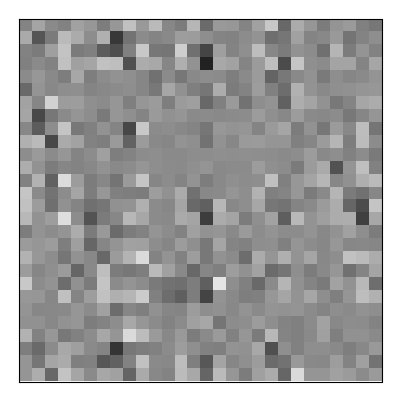}\\
\includegraphics[width=\linewidth]{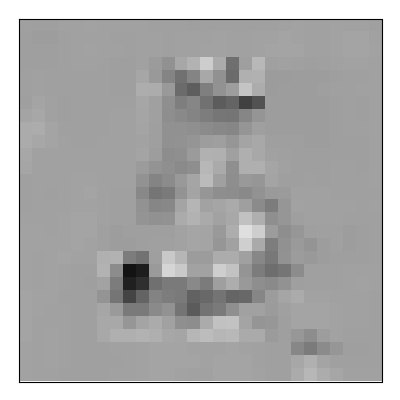}\\
\end{minipage}
\begin{minipage}{0.21\linewidth}
\centering
\tiny
FC VAE\normalsize\\
\includegraphics[width=\linewidth]{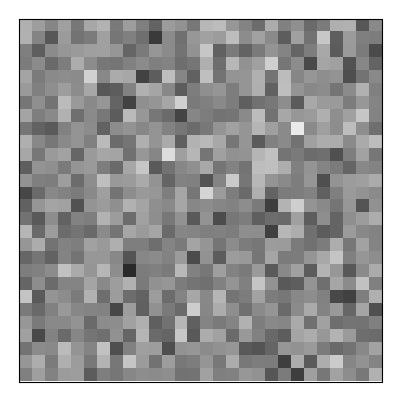}\\
\includegraphics[width=\linewidth]{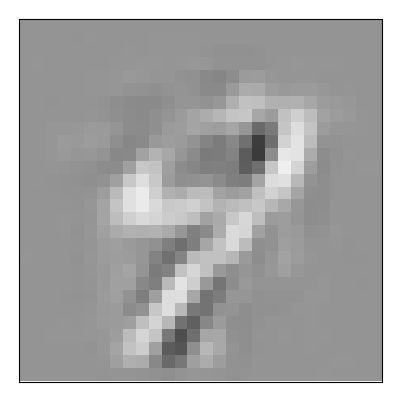}\\
\end{minipage}
\begin{minipage}{0.21\linewidth}
\centering
\tiny
CONV VAE\normalsize\\
\includegraphics[width=\linewidth]{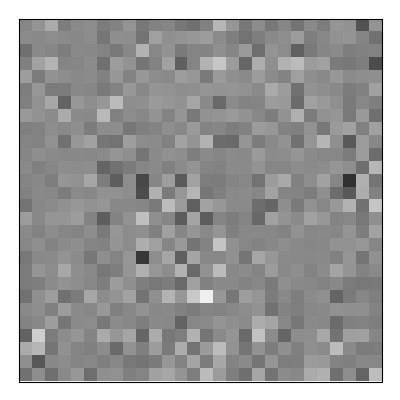}\\
\includegraphics[width=\linewidth]{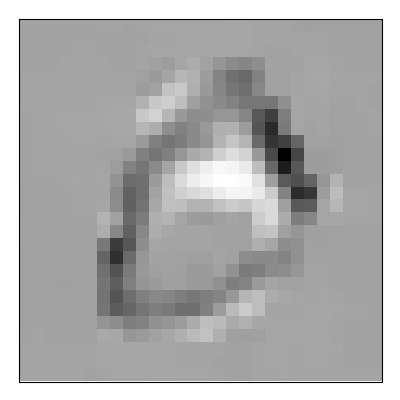}\\
\end{minipage}
\caption{\small{Visualization of a single basis bector $[\bA_{\omega}]_{., k}$ with $\omega$  at initialization and after learning obtained from a region $\omega$ containing the digits $7, 5, 9$, and $0$ respectively, and this for GAN and VAE models made of fully connected or convolutional layer (see Appendix~\ref{appendix:architecture} for details). By visual inspection, one can observe that the depicted basis vector encodes right rotation, cedilla extension, left rotation, and upward translation respectively. 
In addition, we observe how the basis vectors are smoother for VAE-based models which why they tend to generate blurred samples.
}}
\label{fig:basis}
\end{figure}

In this context, we aim to characterize the subspace basis in term of  {\em disentanglement}, i.e.,  the alignment of the basis vectors with respect to each other. 
While there is not a unique definition, a disentangled basis should provide a ``compact'' and interpretable latent representation $\bz$ for the associated $\bx=\bG(\bz)$. In particular, it should ensure that a small perturbation of dimension $d$ of $\bz$ implies a transformation independent from a small perturbation of dimension $d' \not = d$ \cite{schmidhuber1992learning,bengio2013representation}. That is, $\langle \bG(\bz)-\bG(\bz+\epsilon \delta_d),\bG(\bz)-\bG(\bz+\epsilon \delta_{d'})\rangle \approx 0$ with $\delta_d$ a one-hot vector at position $d$ and length $Z$ \cite{kim2018disentangling}. A disentangled representation is thus considered to be most informative as each latent dimension imply a transformation that leaves the others unchanged \cite{bryant1995principal}. For example, rotating an object should not alter its vertical or horizontal translation and vice-versa

\begin{prop}
\label{prop:disentenglement}
A necessary condition for disentanglement is to have ``near orthogonal'' columns as
$\langle [\bA_{\omega}]_{.,i},[\bA_{\omega}]_{., j}\rangle \approx 0,  \forall i,\not = j, \forall \omega \in \Omega$. (Proof in Appendix \ref{proof:disentenglement}.)
\end{prop}
\qq

\small
\begin{table}[t]
    \caption{\small{Depiction of the cosine similarity summed over pairwise different columns of $\bA_{\omega}$. Measure of $0$ means that the basis vectors are orthogonal, improving disentanglement (recall Prop.~\ref{prop:disentenglement}). The first line represent the maximum of this quantity over $10000$ sampled regions $\omega$, the second line represents the average; the std of those quantities is given for $8$ runs.
    We see that training increases disentanglement, and fully connected models offer increased disentanglement as compared to convolutional models.}}
\vspace*{2mm}
\setlength\tabcolsep{2 pt}
    \centering
    \begin{tabular}{|c|c|c|c|c|c|}
\cline{2-5}
\multicolumn{1}{c|}{}&\multicolumn{1}{c|}{{\small FC GAN}} & \multicolumn{1}{c|}{{\small CONV GAN}} & \multicolumn{1}{c|}{{\small FC VAE}} & \multicolumn{1}{c|}{{\small CONV VAE}}\\ \cline{1-5}
\multirow{2}{*}{\rotatebox{90}{init.}}& {\small 8.84 $\pm$ 0.07} & 
{\small 3.2 $\pm$ 0.33}  & {\small 5.23 $\pm$ 0.29} & {\small 3.5 $\pm$ 0.27}  \\  \cline{2-5}
 & {\small 4.41 $\pm$ 0.26} & {\small 1.84 $\pm$ 0.08} & {\small 2.25 $\pm$ 0.08} & {\small 1.74 $\pm$ 0.06} \\ \hline \hline
\multirow{2}{*}{\rotatebox{90}{learn}}& {\small 1.36 $\pm$ .08} & {\small 1.72 $\pm$ 0.07} & {\small 1.32 $\pm$ 0.07} & {\small 1.77 $\pm$ 0.11} \\ \cline{2-5}
& {\small 0.9 $\pm$ 0.03}  & {\small 1.12 $\pm$ 0.03} & {\small 0.89 $\pm$ 0.03} & {\small 1.15 $\pm$ 0.03} \\ \hline
\end{tabular}
    \label{tab:angles}
\end{table}
\normalsize

Figure~\ref{fig:basis} visualizes one of the basis vectors of four different DGNs trained on the MNIST dataset with $S=10$. Interpretability of the transformation encoded by the dimension of the basis vector can be done as well as model comparison such as blurriness of VAE samples that is empirically observed across datasets \cite{zhao2017towards,huang2018introvae}. We also provide in Table~\ref{tab:angles} the value of $\|Q_{\omega} -I\|_2$ with $Q_{\omega} \in [0, 1]^{S \times S}$ the matrix of cosine angles between basis vector of $\bA_{\omega}$ for 10,000 regions sampled randomly and where we report the average over the regions and the maximum. Finally, this process is performed over $8$ runs, the mean and standard deviation are reported in the table. 
We observe that there does not seem to be a difference in the degree of disentanglement different GAN and VAE; however, the topology, fully connected vs.\ convolution, plays an important part, favoring the former.
To visually control the quality of the DGN, randomly generated digits are given in Fig.~\ref{fig:appendix_digits} in the Appendix; we also provide more background on disentanglement in Appendix~\ref{sec:disentangled}.

\subsection{Generated Manifold Curvature}
\label{sec:curvature}

We now study the curvature or angularity of the generated mapping. That is, whenever $\widetilde{S}<D$, the per-region affine subspace of adjacent region are continuous, and joint at the region boundaries with a certain angle that we now characterize.
\begin{defn}
\label{def:adjacent}
Two regions $\omega, \omega'$ are adjacent whenever they share part of their boundary as in $ \overline{\omega} \cap \overline{\omega'} \not = \emptyset$.
\end{defn}
\qq

The angle between adjacent affine subspace is characterized by means of the greatest principal angle \cite{afriat1957orthogonal, bjorck1973numerical} and denote $\theta$.
Denote the per-region projection matrix of the DGN by
\begin{align}
   P(\bA_{\omega}) = \bA_{\omega}(\bA_{\omega}^T\bA_{\omega})^{-1}\bA_{\omega}^T
\end{align}
where $\bA_{\omega}^T\bA_{\omega}\in\mathbb{R}^{S \times S}$ and $P(\bA_{\omega}) \in \mathbb{R}^{D \times D}$. 
We now assume that $\dim(\bG)=Z$ ensuring that $\bA_{\omega}^T\bA_{\omega}$ is invertible.\footnote{The derivation also applies if $\dim(\bG)<Z$ by replacing $\bA_{\omega}$ with $\bA_{\omega}'$ (recall Lemma~\ref{lemma:basis}).}

\begin{thm}
\label{thm:angle}
The angle between adjacent (recall Def.~\ref{def:adjacent}) region mappings $\theta(\bG(\omega), \bG(\omega'))$ is given by
\begin{align}
    \sin\big(\theta(\bG(\omega), \bG(&\omega'))\big)= \| P(\bA_{\omega})- P(\bA_{\omega'})\|_2,\label{eq:angle}
\end{align}
$\forall \omega \in \Omega, \omega' \in \adj(\omega)$. (Proof in Appendix \ref{proof:angle}.)
\end{thm}
\qq

Notice that two special cases of the above theorem emerge. When $S=1$, the angle is given by the cosine similarity between the vectors $\bA_{\omega}$ and $\bA_{\omega'}$ of adjacent regions. When $S=D-1$ the angle is given by the cosine similarity between the normal vectors of the $D-1$ subspace spanned by $\bA_{\omega}$ and $\bA_{\omega'}$ respectively.

\begin{figure}[t]
    \centering
    \begin{minipage}{0.38\linewidth}
    \includegraphics[width=\linewidth]{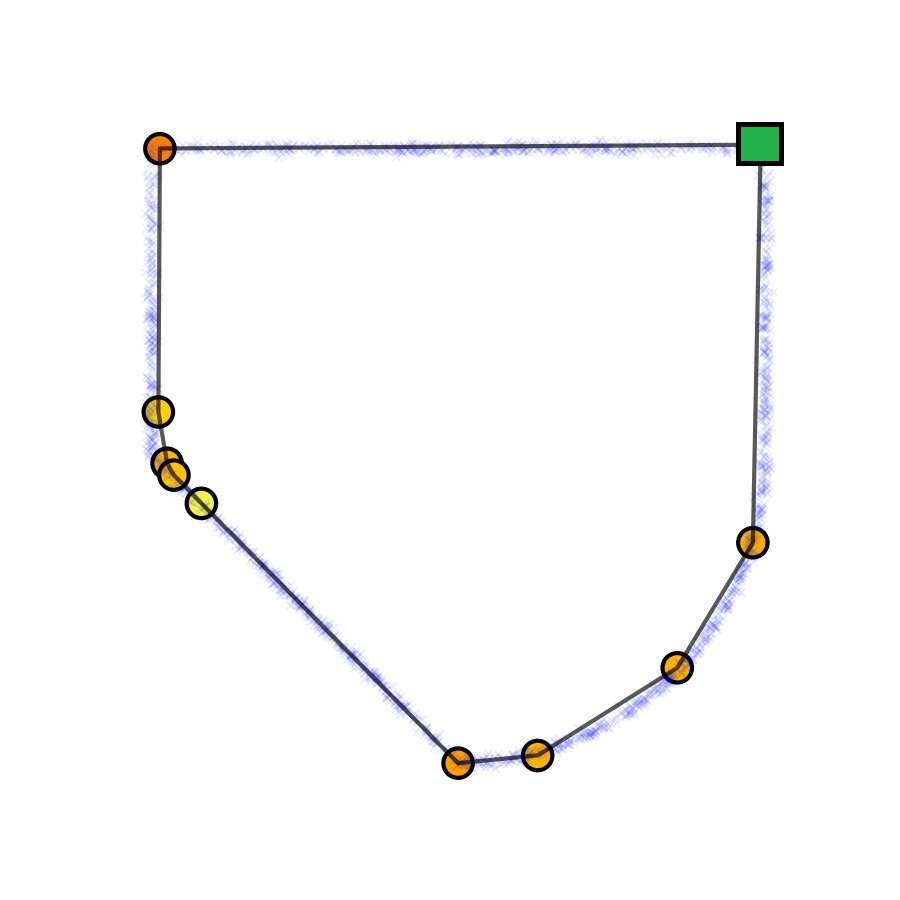}
    \end{minipage}
    \begin{minipage}{0.6\linewidth}
    \caption{
    Piecewise linear continuous 1-D manifold learned by a GAN DGN (in black) from a collection of data points (in blue). 
    The breakpoints between adjacent regions are depicted by dots of color proportional to the angle. The manifold starts at the green box and proceeds clockwise.  Figure~\ref{fig:appendix_1d} in the Appendix contains additional examples.
    }
    \label{fig:angle_2d}
    \end{minipage}
    \vspace*{-5mm}
\end{figure}

We illustrate the angles in a simple case $D=2$ and $Z=1$ in Fig.~\ref{fig:angle_2d}. It can be seen how a DGN with few parameters produces angles mainly at the points of curvature of the manifold. We also provide many additional figures with different training settings in  Fig.~\ref{fig:appendix_angles} in the Appendix as well as repetitions of the same experiment with different random seeds.

\subsection{Application: Angle Distribution of a DGN with Random Weights}

\begin{figure}[t]
    \centering
    \begin{minipage}{0.32\linewidth}
            \centering
            \includegraphics[width=\linewidth]{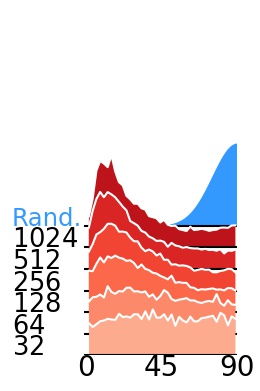}
    \end{minipage}
    \begin{minipage}{0.32\linewidth}
            \centering
            \includegraphics[width=\linewidth]{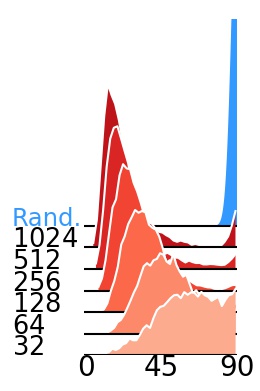}
    \end{minipage}
    \begin{minipage}{0.32\linewidth}
            \centering
            \includegraphics[width=\linewidth]{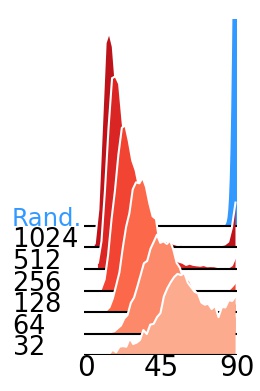}
    \end{minipage}
    \caption{\small 
    Histograms of the largest principal angles for
    DGNs with two hidden layers, $S=16$ and $D=17, D=32, D=64$ respectively and varying width $D_{\ell}$ on the y-axis. Three trends to observe: When the width becomes large, the distribution becomes more bimodal and greatly favors near $0$ angles; when the output space dimension becomes large, there is an increase in the number of angles near orthogonal; the amount of {\em weight sharing} between the parameters $\bA_{\omega}$ and $\bA_{\omega'}$ of adjacent regions $\omega$ and $\omega'$ allow to greatly constrain the angles to be small, as opposed to the distribution of angles between random subspaces \cite{absil2006largest} depicted in blue. Hence despite the large amount of regions, most will be aligned with each other leading to an overall well behave manifold. Additional figures are available in Fig.~\ref{fig:appendix_angles} in the Appendix. \normalsize }
    \label{fig:angles}
\end{figure}

We can use the above result to study the distribution of angles of different DGNs with random weights and study the impact of depth, width, as well $Z$ and $D$, the latent and output dimensions respectively. 
Figure~\ref{fig:angles} summarizes the distribution of angles for several different settings. 

Two key observations emerge.
First, the codes of adjacent regions $\bq(\omega), \bq(\omega')$ share a large number of their values (see Appendix~\ref{appendix:neighbour_codes} for details) implying that most of the DGN parameters are shared in their composition to produce  $\bA_{\omega}$ and $\bA_{\omega'}$. In turn, this {\em weight sharing} correlates adjacent hyperplanes, such that their angles are much smaller than if randomly picked form one another. The random case (in blue in Fig.~\ref{fig:angles}) favors aggressively large angles as opposed to the ones of DGNs.
Second, the distribution moments depend on the ratio $S/D$ rather that those values taken independently. In particular, as this ratio gets smaller, as the angle distribution becomes bi-modal with an emergence of high angles. That is, the manifold is ``flat'' overall except in some parts of the space where high angularity is present. This effect is strengthened with wider DGNs. Notice that this large ratio $S /D$ is the one encountered in practice where it is common to have $S \approx 100$ and $D > 800$.

The above experiment demonstrates the impact of width and latent space dimension into the angularity of the DGN output manifold and how to pick its architecture based on a priori knowledge of the target manifold. Under the often-made assumptions that the weights of overparametrized DGN do not move far from their initialization during training \cite{li2018learning}, these results also hint at the distribution of angles after training.

\section{Density on the Generated Manifold}

The study of DGNs would not be complete without considering that the latent space is equipped with a density distribution $\bp_{\bz}$ from which $\bz$ are sampled in turn leading to sampling of $\bG(\bz)$. 
Thus, we now study how this density is spread over the output space, covering the generated manifold and highlighting some key properties such as density conentration, entropy computation and training instabilities.

\subsection{Analytical Output Density}
\label{sec:output_density}

Given a distribution $\bp_{\bz}$ over the latent space, we can explicitly compute the output distribution after the application of $\bG$, which lead to an intuitive result exploiting the piecewise affine property of the generator.

\begin{lemma}
\label{lemma:volume}
Denote by $\sigma_i(\bA_{\omega})$ the $i^{\rm th}$ singular value of $\bA_{\omega}$.
Then, the volume of a region $\omega \in \Omega$ denoted by $\mu(\omega)$ is related to the volume of $\bG(\omega)$ by
\begin{align*}
    \mu(\bG(\omega)) = \sqrt{\det(\bA_{\omega}^T\bA_{\omega})}\mu(\omega) = \prod_{i: \sigma_i(\bA_{\omega}) >0} \hspace{-0.4cm}\sigma_i(\bA_{\omega})\mu(\omega).
\end{align*}
(Proof in Appendix \ref{proof:volume}.)
\end{lemma}
\qq

\begin{thm}
\label{thm:general_density}
The generator probability density  $p_{\bG}(\bx)$ given $\bp_{\bz}$ and a injective generator $\bG$ with per-region inverse $\bG^{-1}_{\omega}$ from (\ref{eq:region_inverse}) is given by
\begin{align}
    p_{\bG}(\bx) = \sum_{\omega \in \Omega} \frac{\bp_{\bz}\left(\bG^{-1}_{\omega}(\bx)\right)}{\sqrt{\det(\bA_{\omega}^T\bA_{\omega})}}\Indic_{\{\bx \in \bG(\omega)\}}.\label{eq:general_density}
\end{align}
(Proof in Appendix \ref{proof:general_density}.)
\end{thm}
\qq

That is, the distribution obtained in the output space naturally corresponds to a piecewise affine transformation of the original latent space distribution, weighted by the change in volume of the per-region mappings.

From the analytical derivation of the generator density distribution, we obtain its differential entropy, i.e., the Shannon entropy for continuous distributions.

\begin{cor}
\label{cor:entropy}
The differential entropy of the output distribution $\bp_{\bG}$ of the DGN is given by
\begin{align*}
    E(\bp_{\bG})=E(\bp_{\bz})+\sum_{\omega \in \Omega} P(\bz \in \omega)\log(\sqrt{\det(\bA_{\omega}^T\bA_{\omega})}).
\end{align*}
(Proof in Appendix \ref{proof:entropy}.)
\end{cor}
\qq

As the result, the differential entropy of the output distribution $\bp_{\bG}$ corresponds to the differential entropy of the latent distribution $\bp_{\bz}$ plus a convex combination of the per-region volume change.
Two results emerge directly. First, it is possible to optimize the latent distribution $\bp_{\bz}$ to better fit the target distribution entropy as been done for example in \cite{ben2018gaussian}. Second, whenever this distribution is fixed, any gap between the latent and output distribution entropy imply the need for high change in volumes between $\omega$ and $\bG(\omega)$.

{\bf Gaussian Case.}
We now demonstrate the use of the above derivation by considering practical examples for which we are able to gain ingights into the DGN data modeling and generation. First, consider that the latent distribution is set as  $\bz \sim \mathcal{N}(0,1)$ we obtain the following result directly from Thm.~\ref{thm:general_density}.

\begin{cor}
\label{cor:gaussian_density}
The generator density distribution $p_{\bG}(\bx)$ given $\bz \sim \mathcal{N}(\mathbf{0},I)$ is
\begin{align*}
    p_{\bG}(\bx) = \sum_{\omega \in \Omega}\frac{e^{-\frac{1}{2}(\bx-\bb_{\omega})^T(\bA_{\omega}^+)^T\bA_{\omega}^+(\bx-\bb_{\omega})} }{\sqrt{(2\pi)^{S}\det(\bA_{\omega}^T\bA_{\omega})}}\Indic_{\{\bx \in \bG(\omega)\}}.
\end{align*}
(Proof in Appendix \ref{proof:gaussian_density}.)
\end{cor}
\qq

The above formula is reminiscent of Kernel Density Estimation (KDE) \cite{rosenblatt1956remarks} and in particular adaptive KDE \cite{breiman1977variable}, where a partitioning of the data manifold is performed on each cell ($\omega$ in our case) different kernel parameters are used.

{\bf Uniform Case.}
We now turn into the uniform latent distribution case.
Consider the following question: {\em Suppose we start from a uniform distribution $\bz \sim \mathcal{U}(0,1)$ on the hypercube in $\mathbb{R}^S$, will the samples be uniformly distributed on the manifold of $\bG$?}

\begin{prop}
Given a uniform latent distribution $\bv \sim \mathcal{U}(0,1)$, the sampling of the manifold $\bG(\text{supp}(\bp_{\bz}))$ will be uniform iff $\det(\bA^T_{\omega}\bA_{\omega})=c , \forall \omega: \omega \cap \text{supp}(\bp_{\bz}) \not = \emptyset, c>0$.
\end{prop}
\qq

\subsection{Generative Deep Network Likelihood and Normalizing Flows}
\label{sec:nmf}

Note from Thm.~\ref{thm:general_density} that we obtain an explicit density distribution. One possibility for learning thus corresponds to minimizing the negative log-likelihood (NLL) between the generator output distribution and the data. Recall from Thm.~\ref{thm:general_density} that $\sqrt{\det{((\bA^{+}_{\omega})^T\bA^{+}_{\omega})}}=(\sqrt{\det{(\bA_{\omega}^T\bA_{\omega})}})^{-1}$; thus we can write the log density from (\ref{eq:general_density}) over a sample $\bx_n$ as 
$
    \mathcal{L}(\bx_n) \hspace{-0.05cm}=\hspace{-0.05cm} \log (\bp_{\bz}(\bG^{-1}(\bx_n))) \hspace{-0.05cm}+\hspace{-0.05cm} \log(\sqrt{\det(J(\bx_n))})$,
where $J(\bx_n)=J_{\bG^{-1}}(\bx_n)^TJ_{\bG^{-1}}(\bx_n)$, $J$ the Jacobian operator.
Learning the weights of the DGN by minimization of the NLL given by $-\sum_{n=1}^N\mathcal{L}(\bx_n)$, corresponds to the normalizing flow model. The practical difference between this formulation and most NF models comes from having either a mapping from $\bx \mapsto \bz$ (NF) or from $\bz \mapsto \bx$ (DGN case). This change only impacts the speed to either sample points or to compute the probability of observations. In fact, the forward pass of a DGN is easily obtained as opposed to its inverse requiring a search over the codes $\bq$ itself requiring some optimization. Thus, the DGN formulation will have  inefficient training (slow to compute the likelihood) but fast sampling while NMFs will have efficient training but inefficient sampling. 

\subsection{On the Difficulty of Generating Low entropy/Multimodal Distributions}
\label{sec:difficulty}

We conclude this section with the study of the instabilities encountered when training DGNs on multimodal densities or other atypical cases.

\begin{figure}[t]
    \centering
    \begin{minipage}{0.32\linewidth}
    \includegraphics[width=\linewidth]{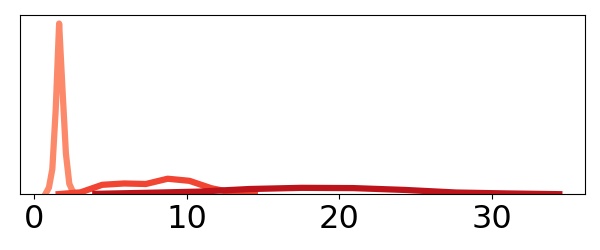}\\
    {\scriptsize $\sigma_1=0,\sigma_2\in \{1,2, 3\}$}
    \end{minipage}
    \begin{minipage}{0.32\linewidth}
    \includegraphics[width=\linewidth]{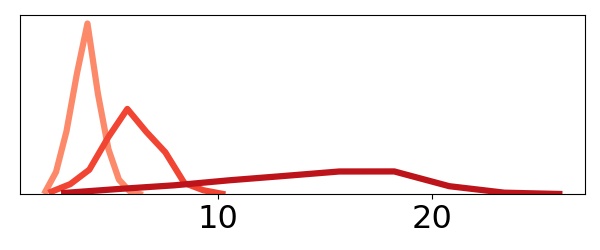}\\
    {\scriptsize $\sigma_1=1,\sigma_2\in \{1,2, 3\}$}
    \end{minipage}
    \begin{minipage}{0.32\linewidth}
    \includegraphics[width=\linewidth]{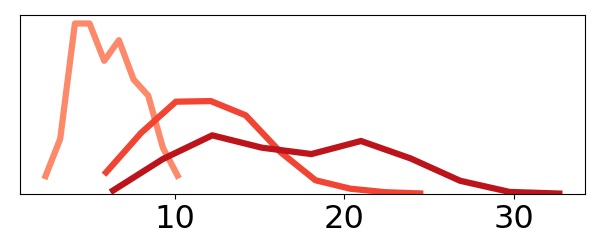}\\
    {\scriptsize $\sigma_1=2,\sigma_2\in \{1,2, 3\}$}
    \end{minipage}
    \caption{\small{Distribution of  $\log(\sqrt{\det(\bA_{\omega}^T\bA_{\omega})})$ for $2000$ regions $\omega$ with a DGN with {\scriptsize $L=3, S=6, D=10$} and weights initialized with Xavier; then, half of the weights' coefficients (picked randomly) are rescaled by $\sigma_1$ and the other half by $\sigma_2$. We observe that greater variance of the weights increase the spread of the log-determinants and increase the mean of the distribution.}}
    \label{fig:example_dets}
\end{figure}

\begin{figure}[t]
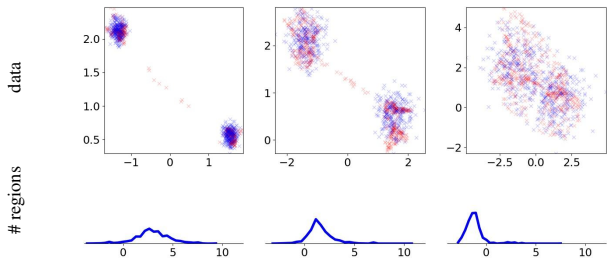

\begin{minipage}{0.1\linewidth}
\rotatebox{90}{{\scriptsize  \# regions} \hspace{0.8cm} {\scriptsize data}}
\end{minipage}
\begin{minipage}{0.88\linewidth}
\foreach \std in {3,4,5}{
    \begin{minipage}{0.32\linewidth}
    \foreach \run in {3}{
        \includegraphics[width=\linewidth]{determinant/data_2d_std\std_run\run.jpg}\\
        \includegraphics[width=\linewidth]{determinant/deter_2d_std\std_run\run.jpg}\\
    }
    \end{minipage}}
\end{minipage}
\vspace*{-4mm}
    \caption{\small{Distribution of the log-determinant of the per-region mappings for different true distributions: two Gaussian with increasing standard deviation (left to right). We observe how the concentration and inter-mode distance impacts greatly the distribution of the log-determinant to allow the generator to fit the distribution, this in turns increases the variance of the weights $\bW_{\ell}$.}}
    \label{fig:blobs}
\end{figure}

\begin{figure}[t]
    \centering
    \begin{minipage}{0.32\linewidth}
    \includegraphics[width=\linewidth]{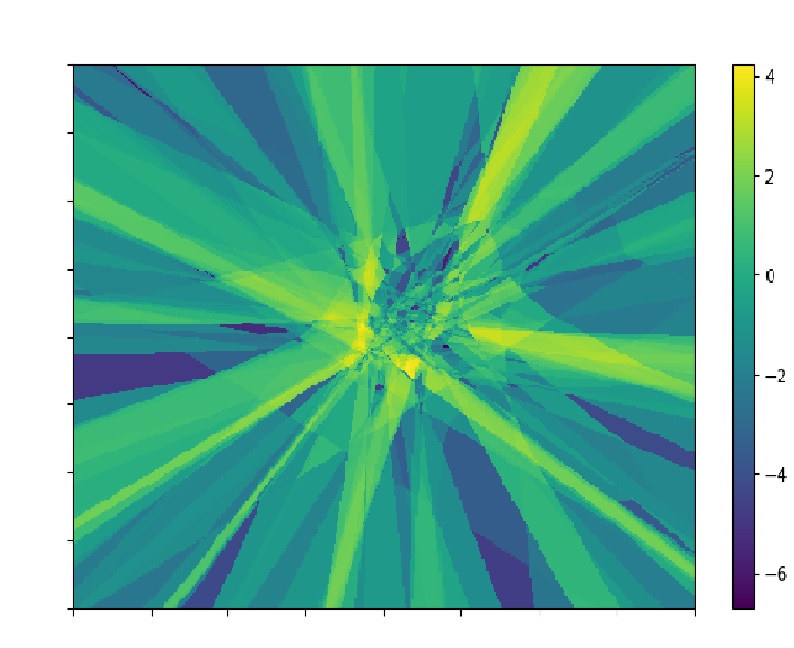}
    \end{minipage}
    \begin{minipage}{0.32\linewidth}
    \includegraphics[width=\linewidth]{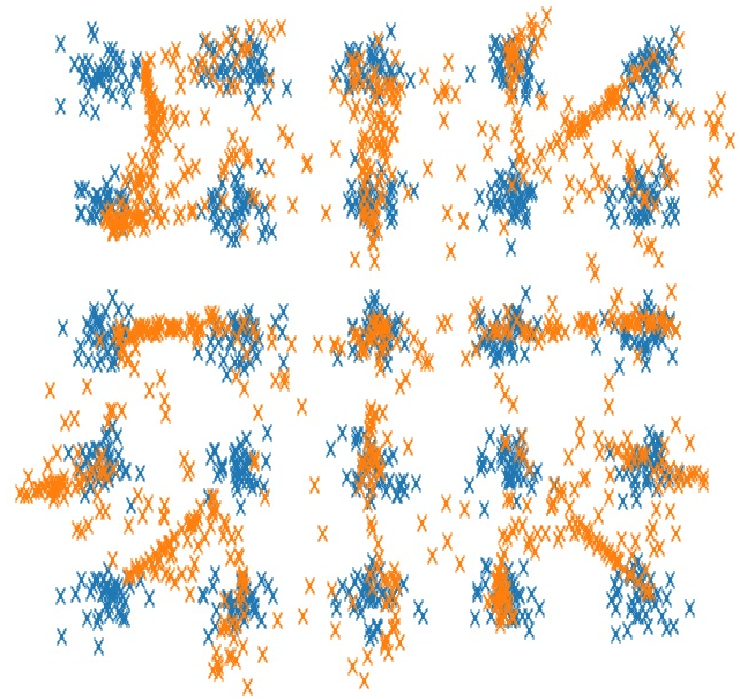}
    \end{minipage}
    \begin{minipage}{0.32\linewidth}
    \includegraphics[width=\linewidth]{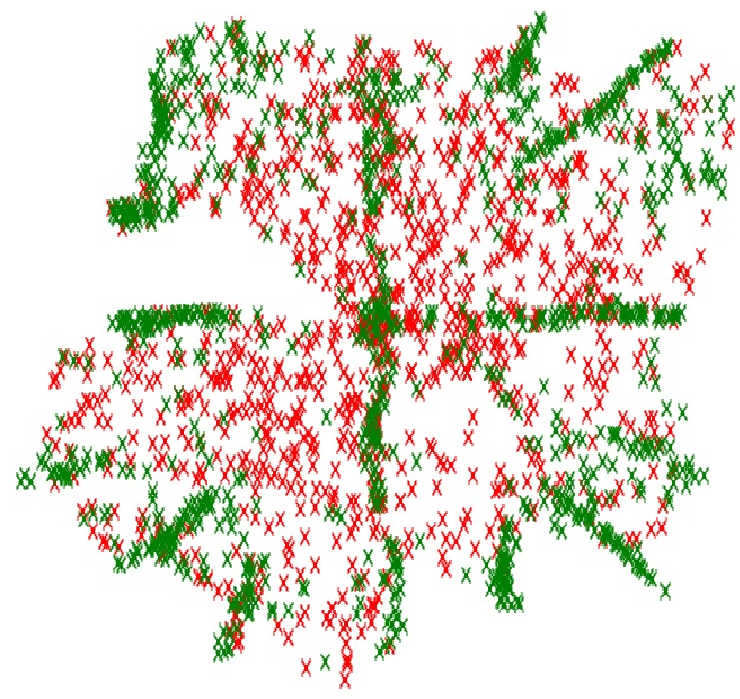}
    \end{minipage}
    \caption{\small{Example of a GAN DGN trained on a mixture of $25$ Gaussians. On the {\bf left} is depicted the latent space per-region log-determinant (color coded) with values ranging from $-6$ to $4$ in log-scale ($0.002$ to $55$ in linear scale). On the {\bf middle} are depicted the true sample (blue) and generated ones (oranges), and on the {\bf right} are depicted points sampled from the generator from regions with low determinant (green) and large determinant (red). We can observe that this failure case (poor approximation of the true distribution) is due to the continuous properties of MASO DGNs, which makes the generator move continuously between modes while not being able to reduce enough the sampling probability $\bp_{\bG}$ in between the modes. Additional examples are contained in Fig.~\ref{fig:appendix_blobs}}}
    \label{fig:det_cases}
\end{figure}

We demonstrated in Thm.~\ref{thm:general_density} and Cor.~\ref{cor:entropy}  that the product of the nonzero singular values of $\bA_{\omega}$ plays the central role to concentrate or disperse the density on $\bG(\omega)$.
Let now consider a simple case of mixture of Gaussians. It becomes clear that the standard deviation of the modes and the inter-mode distances will put constraints on the singular values of the slope matrix $\bA_{\omega}$. However, imposing a large variance in the singular values of $\bA_{\omega}$ for different regions $\omega$ directly stress the parameters $\bW_{\ell}$ as they compose the slope matrix. This is highlighted in Fig.~\ref{fig:example_dets} where we depict the distribution of the log-determinants for different bimodal distribution for the weights $\bW_{\ell}$ showing the correlation between the variance of those parameters and the variance of log-determinant over different regions.

We also illustrate the variation of the learned generator log-determinant distribution across regions in Fig.~\ref{fig:blobs}, where we trained a GAN DGN on two Gaussians for different scalings. This further highlights the importance of the distribution of the determinants that is reachable by a DGN which depends on the architecture and parameter space. In conclusion, for multimodal and low entropy distribution, the required log-determinant for the DGN to approximate the true distribution goes against some standard regularization techniques such as Tikhonov, which (recall Fig.~\ref{fig:example_dets}) pushes the generator output density $\bp_{\bG}$ to be more uniform  with higher entropy.


\clearpage
\appendix
\onecolumn
\begin{center}
\large{
    \textsc{Supplementary Material}}
\end{center}

\section{Extra Figures}
All the below pictures have been compressed to be uploaded on Arxiv.
\subsection{Angles Histogram}

\begin{figure}[H]
    \centering
    \begin{minipage}{0.16\linewidth}
            \centering
            \small{S=2, D=3} \\
            \includegraphics[width=\linewidth]{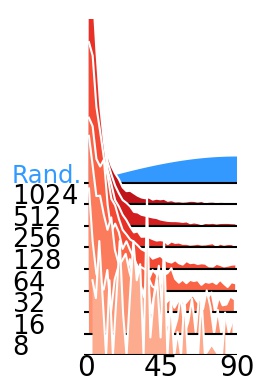}\\
            \includegraphics[width=\linewidth]{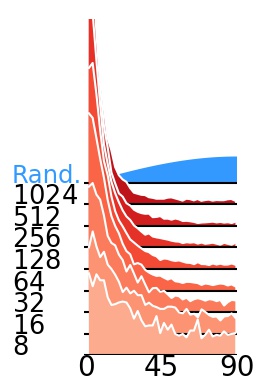}\\
            \small{S=8, D=9} \\
            \includegraphics[width=\linewidth]{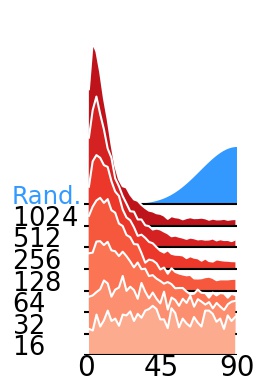}\\
            \includegraphics[width=\linewidth]{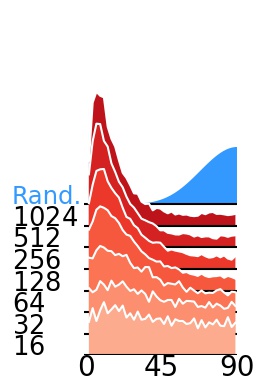}
    \end{minipage}
    \begin{minipage}{0.16\linewidth}
    \centering
            \small{S=2, D=4} \\
            \includegraphics[width=\linewidth]{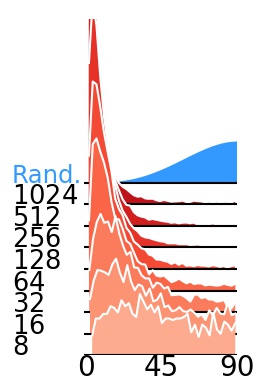}\\
            \includegraphics[width=\linewidth]{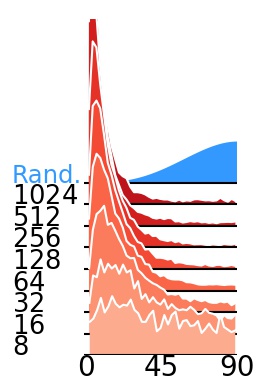}\\
            \small{S=8, D=16} \\
            \includegraphics[width=\linewidth]{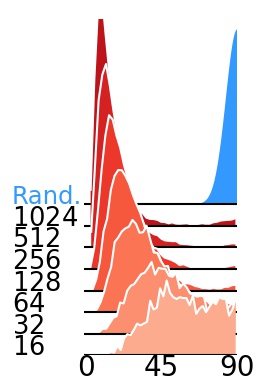}\\
            \includegraphics[width=\linewidth]{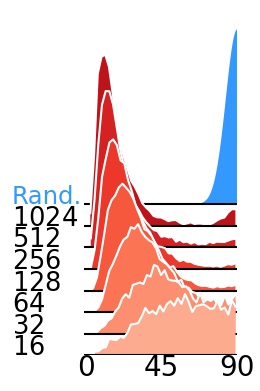}
    \end{minipage}
    \begin{minipage}{0.16\linewidth}
            \centering
            \small{S=2, D=8} \\
            \includegraphics[width=\linewidth]{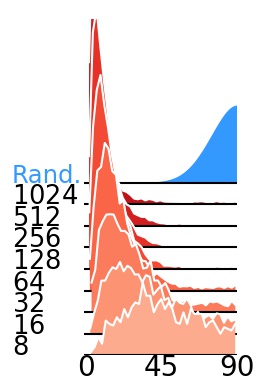}\\
            \includegraphics[width=\linewidth]{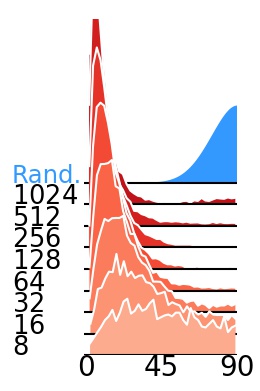}
            \\
            \small{S=8, D=32} \\
            \includegraphics[width=\linewidth]{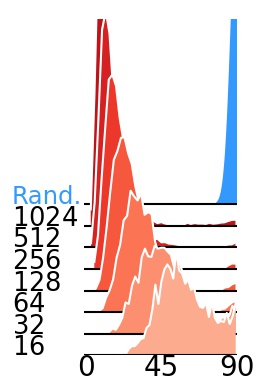}\\
            \includegraphics[width=\linewidth]{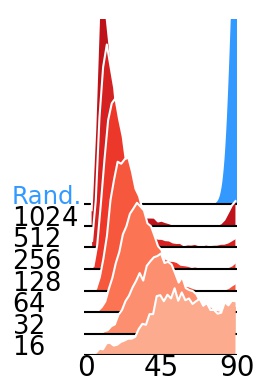}
    \end{minipage}
    \begin{minipage}{0.16\linewidth}
    \centering
            \small{S=4, D=5}\\
            \includegraphics[width=\linewidth]{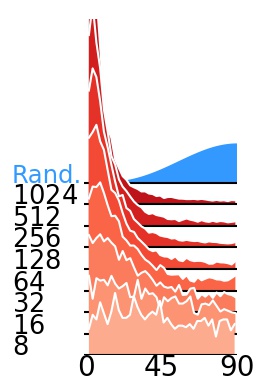}\\
            \includegraphics[width=\linewidth]{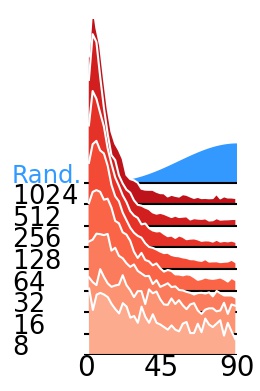}\\
            \small{S=16, D=17} \\
            \includegraphics[width=\linewidth]{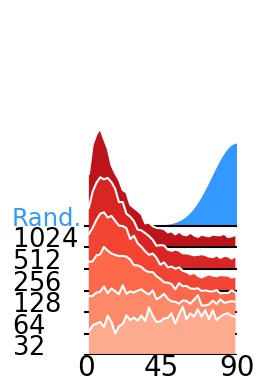}\\
            \includegraphics[width=\linewidth]{hist_angle/angles_histo_2_16_17.jpg}
    \end{minipage}
    \begin{minipage}{0.16\linewidth}
    \centering
            \small{S=4, D=8}\\
            \includegraphics[width=\linewidth]{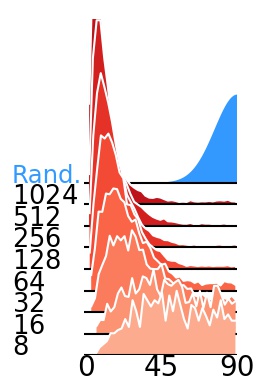}\\
            \includegraphics[width=\linewidth]{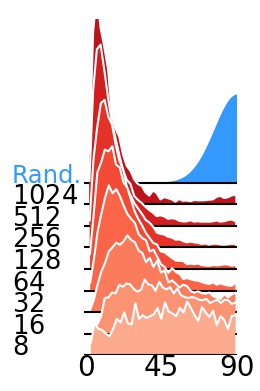}\\
            \small{S=16, D=32} \\
            \includegraphics[width=\linewidth]{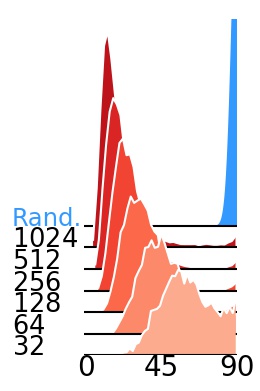}\\
            \includegraphics[width=\linewidth]{hist_angle/angles_histo_2_16_32.jpg}
    \end{minipage}
    \begin{minipage}{0.16\linewidth}
    \centering
            \small{S=4, D=16}\\
            \includegraphics[width=\linewidth]{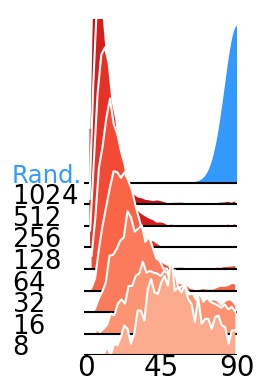}\\
            \includegraphics[width=\linewidth]{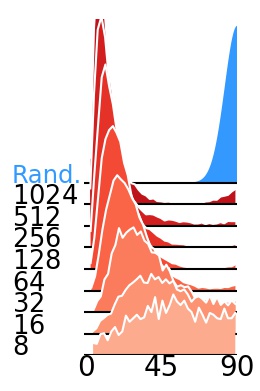}\\
            \small{S=16, D=64} \\
            \includegraphics[width=\linewidth]{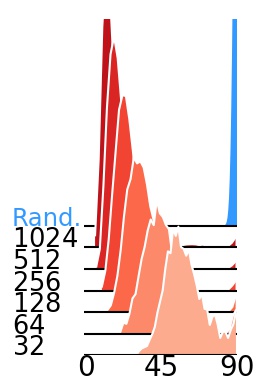}\\
            \includegraphics[width=\linewidth]{hist_angle/angles_histo_2_16_64.jpg}
    \end{minipage}
    \caption{Reproduction of Fig.~\ref{fig:angles}.
    Histograms of the largest principal angles for
    DGNs with one hidden layer (first two rows) and two hidden layers (last two rows). In each case the latent space dimension and width of the hidden layers is in the top of the column. The observations reinforce the claims on the role of width and $S$ versus $D$ dimensions.    
    }
    \label{fig:appendix_angles}
\end{figure}

\subsection{Angles Manifold}

\begin{figure}[H]
\foreach \K in {6, 8, 16, 32}{
    \begin{minipage}{0.24\linewidth}
    \foreach \run in {0, 1, 6, 7, 9}{
        \includegraphics[width=\linewidth]{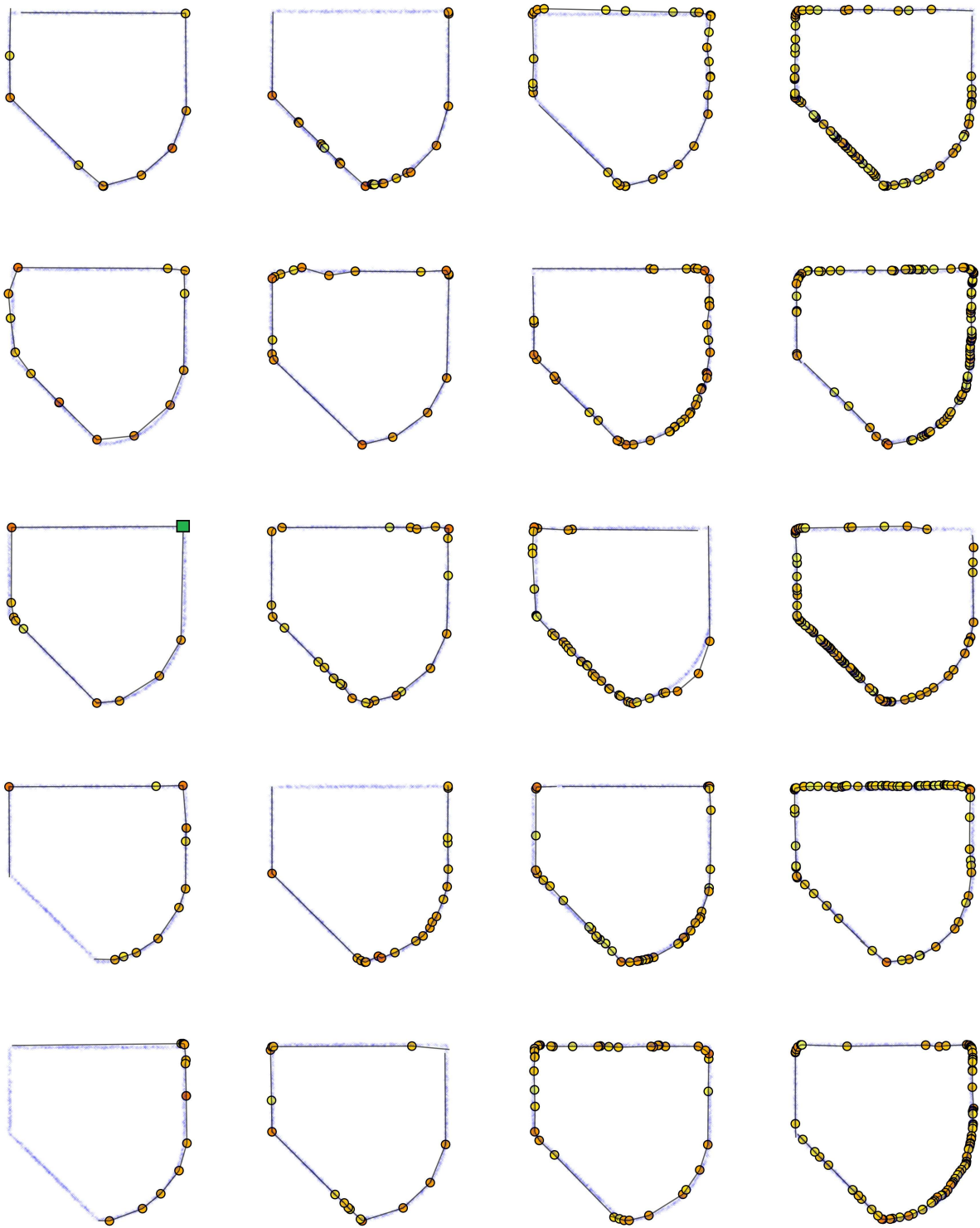}\\
    }
    \end{minipage}}
    \caption{Reproduction of Fig.~\ref{fig:angle_2d} for various GDNs topologies. The columns represent different widths $D_{\ell} \in \{6, 8, 16, 32\}$ and the rows correspond to repetition of the learning for different random initializations of the GDNs for consecutive seeds.}
    \label{fig:appendix_1d}
\end{figure}

\subsection{More on MNIST Disentanglement}

\begin{center}
\begin{figure}[H]
    \centering
    \includegraphics[width=0.7\linewidth]{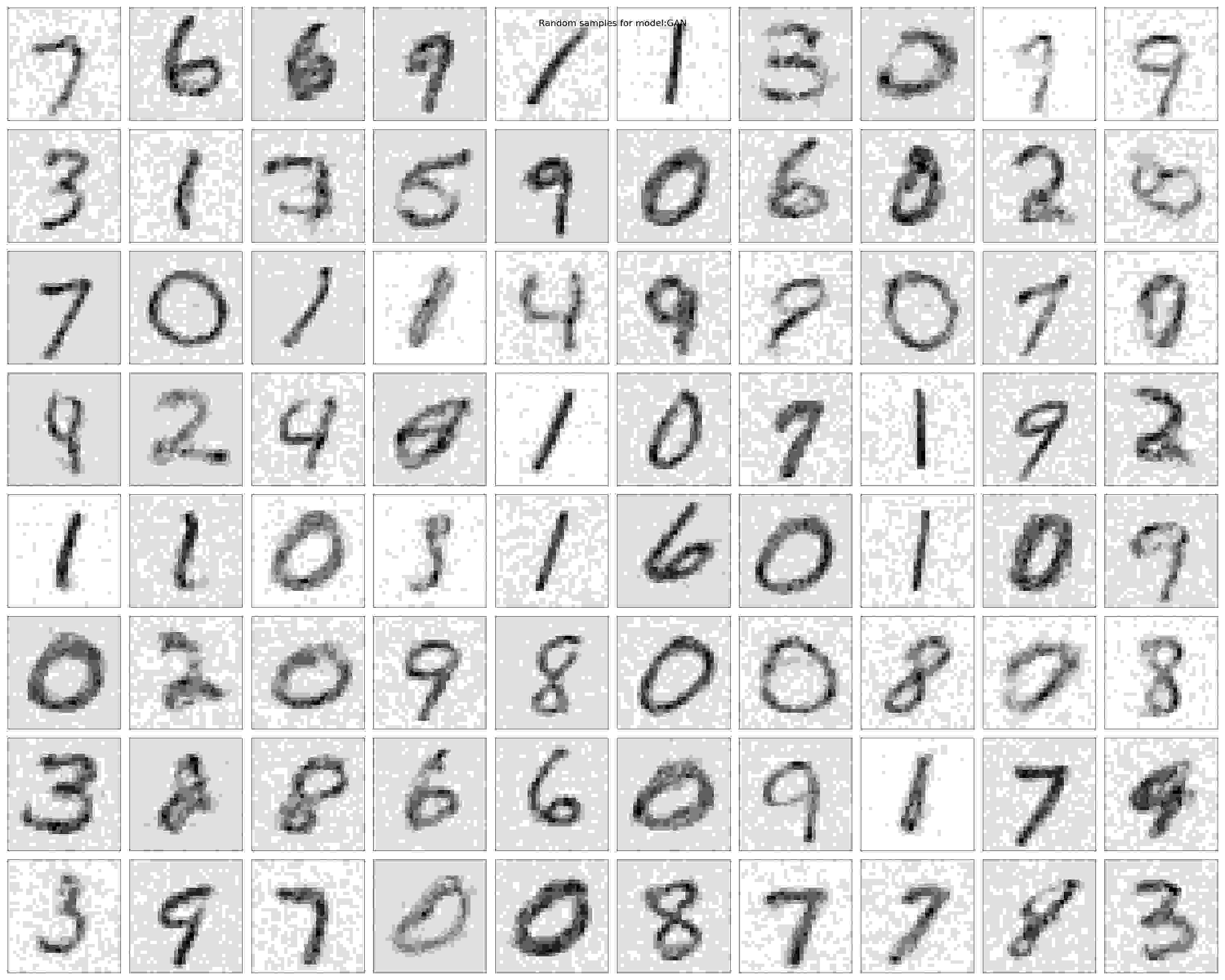}\\
        \includegraphics[width=0.7\linewidth]{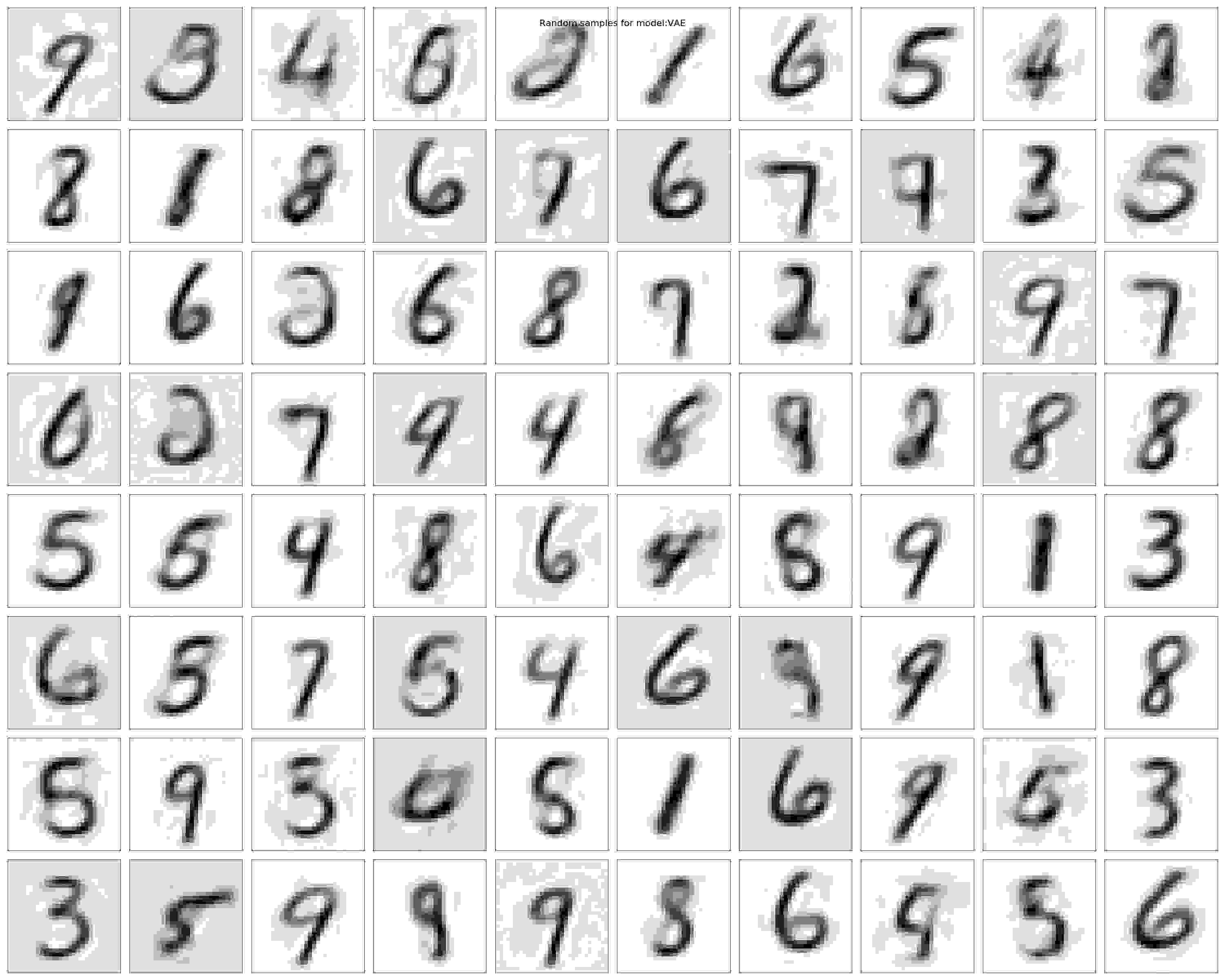}    
    \caption{Randomly generated digits from the trained GAN (top) and trained VAE(bottom) models for the experiment from Fig.~\ref{fig:basis}. Each row represents a model that was training on a different random initialization (8 runs in total) which produced the result in Table~\ref{tab:angles}.}
    \label{fig:appendix_digits}
\end{figure}

\begin{figure}[H]
    \centering
    \includegraphics[width=0.7\linewidth]{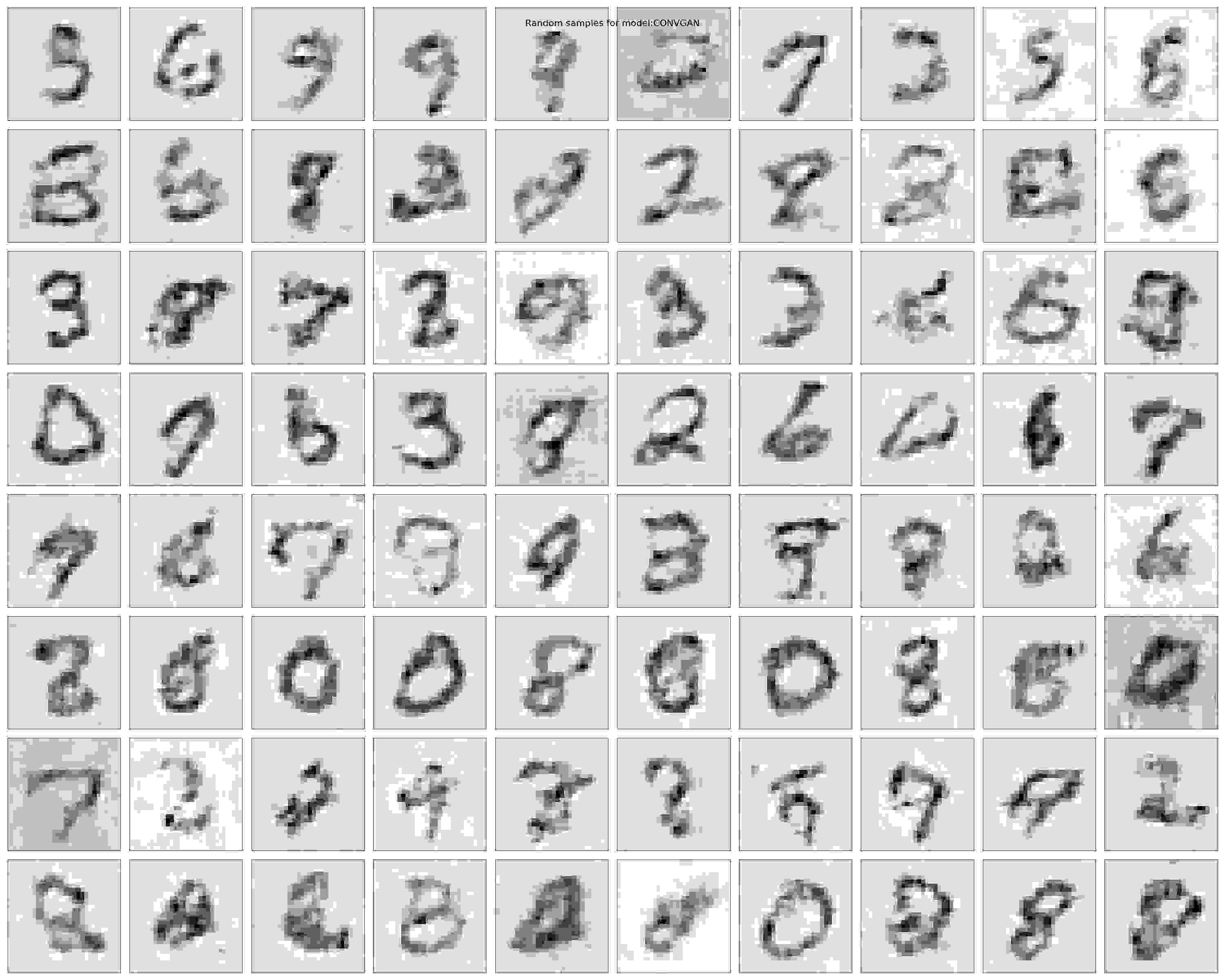}\\
        \includegraphics[width=0.7\linewidth]{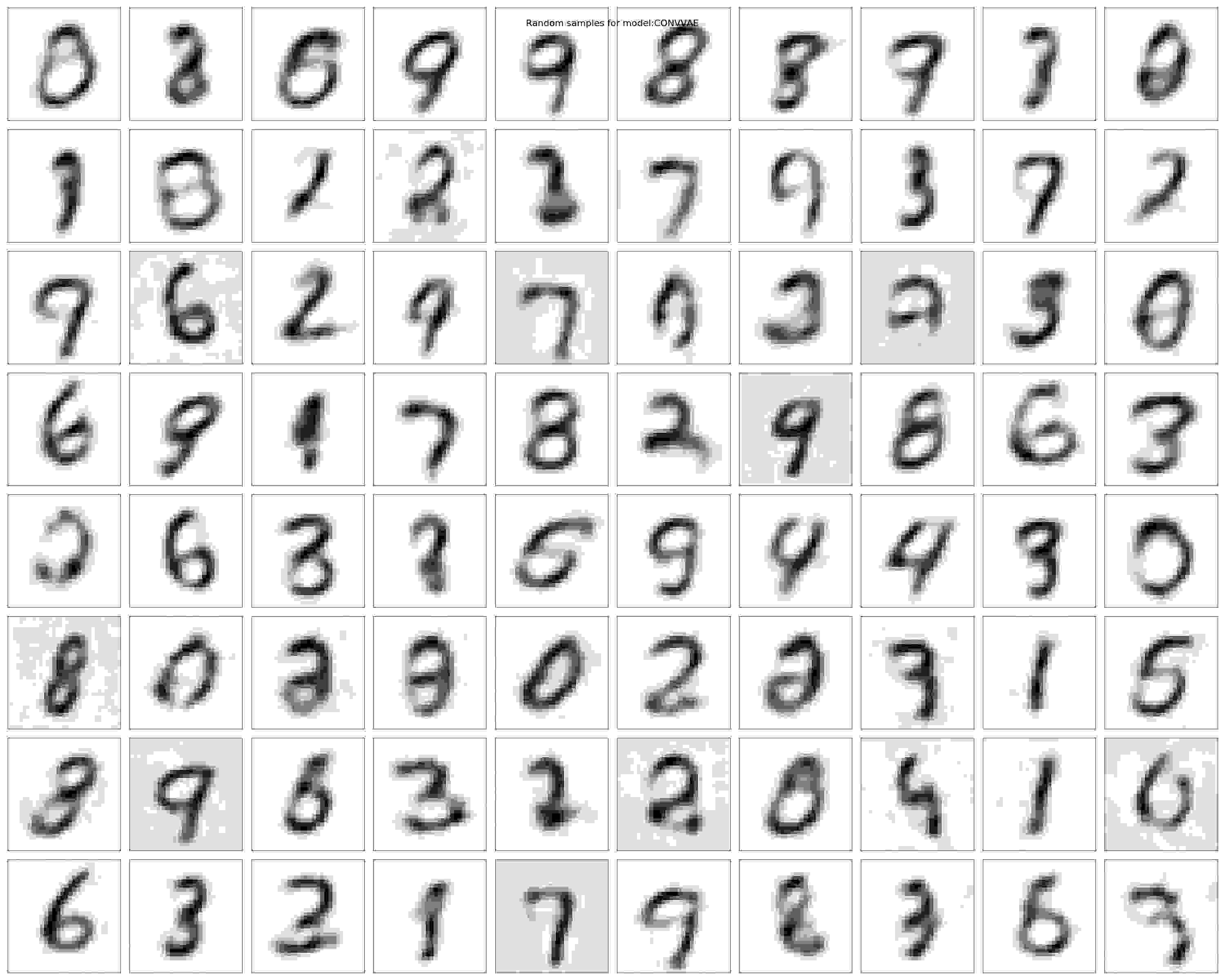}    
    \caption{Randomly generated digits from the trained CONV GAN (top) and trained CONV VAE(bottom) models for the experiment from Fig.~\ref{fig:basis}. Each row represents a model that was training on a different random initialization (8 runs in total) which produced the result in Table~\ref{tab:angles}.}
    \label{fig:appendix_digits}
\end{figure}
\end{center}

\subsection{More on Determinant Figures}
\begin{center}
\begin{figure}[H]
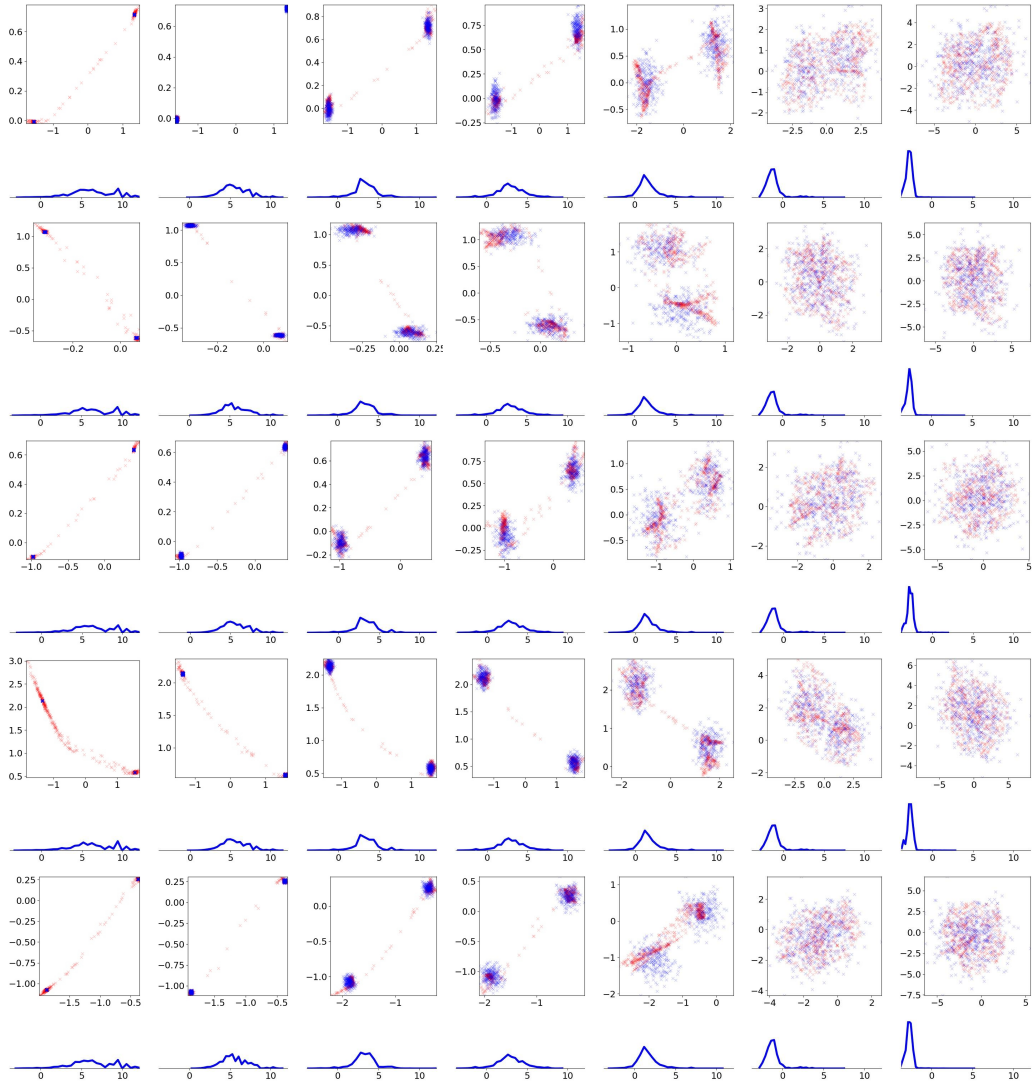

\centering
\foreach \std in {0, 1, 2, 3, 4, 5, 6}{
    \begin{minipage}{0.11\linewidth}
    \foreach \run in {0, 1, 2, 3, 4}{
        \includegraphics[width=\linewidth]{determinant/data_2d_std\std_run\run.jpg}\\
        \includegraphics[width=\linewidth]{determinant/deter_2d_std\std_run\run.jpg}\\
    }
    \end{minipage}}
    \caption{Reproduction of Fig.~\ref{fig:blobs} for multiple standard deviations and multiple random seeds. Each column represent a different standard deviation of the two Gaussians $\sigma \in \{ 0.002, 0.01, 0.05, 0.1, 0.3, 1, 2 \}$ and each row is a run with a different seed. As can be seen in all cases (except when lack of convergence) the distribution of the determinants support the claim and relate with the Entropy of the true distribution (blue points).}
    \label{fig:appendix_blobs}
\end{figure}
\end{center}

\clearpage

\section{Architecture Details}
\label{appendix:architecture}

We describe the used models below. The Dense(T) represents a fully connected layer with $T$ units (activation function not included). The Conv2D(I, J, K) represent $I$ filters of spatial shape $(J,K)$ and the input dilation and padding follow the standard definition. For the VAE models the encoder is given below and for the GAN models the discriminator is given below as well. FC GAN model means that the FC generator is used in conjonction with the discriminator, the CONV GAN means that the CONV generator is used in conjonction with the discriminator and similarly for the VAE case.

\begin{center}
\begin{tabular}{|l|l|l|l|}\hline
    FC generator & CONV generator & Encoder & Discriminator\\ \hline
    Dense(256) &  Dense(256) & Dense(512) & Dense(1024)\\
    leaky ReLU & leaky ReLU & Dropout(0.3) & Dropout(0.3)\\
    Dense(512) & Dense(8 * 6 * 6) & leaky ReLU & leaky ReLU\\
    leaky ReLU & leaky ReLU & Dense(256) & Dense(512)\\
    Dense(1024)& Reshape(8, 6, 6) & leaky ReLU & Dropout(0.3)\\
    leaky ReLU & Conv2D(8, 3, 3, inputdilation=2, pad=same) & Dense(2*S) & leaky ReLU\\
    Dense(28*28) & leaky ReLU & &Dense(256)\\
                 & Conv2D(8, 4, 4, inputdilation=3, pad=valid) && Dropout(0.3)\\
                 & Reshape(28*28) & & leaky ReLU\\
                 &&&Dense(2)\\ \hline
\end{tabular}     
\end{center}
all the training procedures employ the Adam optimizer with a learning of $0.0001$ which stays constant until training completion. In all cases trainig is done on $300$ epochs, an epoch consisting of viewing the entire image training set once.

\section{Proofs}

\subsection{Proof of Proposition \ref{prop:input_region_convexity}}
\label{proof:prop:input_region_convexity}

\begin{proof}
The result is a direct application of Corollary~3 in \cite{balestriero2019geometry} adapted to GDNs (and not classification based DNs). The input regions are proven to be convex polytopes. Then by linearity of the per region mapping, conexity is preserved and with form given by (\ref{eq:region_mapping}).
\end{proof}

\subsection{Proof of Lemma~\ref{lemma:upper_bound}}
\label{proof:upper_bound}

\begin{proof}
First recall the standard result that
\begin{align*}
    \rank(AB)\leq \min(\rank(A),\rank(B)),
\end{align*}
for any matrix $A\in\mathbb{R}^{N \times K}$ and $B \in \mathbb{R}^{K \times D}$ (see for example \cite{banerjee2014linear} chapter 5).
Now, noticing that $\min(\min(a,b),\min(c,d))=\min(a,b,c,d)$ leads to the desired result by unrolling the product of matrices that make up the $\bA_{\omega}$ matrix to obtain the desired result. 
\end{proof}

\subsection{Proof of Proposition~\ref{prop:bijective}}
\label{proof:bijective}

\begin{proof}
First notice that there can only be two major cases. First for the dimension of the affinely mapped region $\bG(\omega)$ to be $S$ or to be smaller than $S$.
Let first consider the bijective case. For the GDN to be bijective on the region we need a one-to-one mapping from $\omega$ to $\bG(\omega)$. If the dimension of the subsapce $\bG(\omega)$ is $S$, then it means that the matrix $\bA_{\omega}$ is full-rank, with rank $S$. In turn, this means that the columns of the matrix are linearly independent. This implies bijectivity on the region as each point in $\omega$ is mapped to an unique point in $\bG(\omega)$ and vice-versa. The surjectivity is direct as if the dimension is smaller than $S$, then the matrix $\bA_{\omega}$ is not full-rank and all the points in the region $\omega$ that leave in the kernel of the matrix (lifted with the bias $\bb_{\omega}$) will be mapped to the same output points. This means that there exists different points in $\omega$ s.t. they are mapped to the same point in $\bG(\omega)$ which gives surjectivity.

For global bijectivity, we need an additional condition. In fact, to ensure that the entire GDN preserves a one-to-one mapping, we need per region bijectivity coupled with the fact that the mapping for different region do not intersect. In fact, we know look at bijectivity between $\text{supp}(\bp_{\bz})$ and $\bG(\text{supp}(\bp_{\bz}))$. Thus if the regions do not intersection after affine projections then there does not exist different latent vectors $\bz$ and $\bz'$ that would be mapped to the same output point. Yet because we have bijectivity on between $\omega$ and $\bG(\omega), \forall \omega$ it means that each point in $\text{supp}(\bp_{\bz})$ is mapped to an unique point in $\bG(\text{supp}(\bp_{\bz}))$ which gives global bijectivity.

\end{proof}

\subsection{Proof of Proposition~\ref{prop:mixture}}
\label{proof:dim_dropout_range}

\begin{proof}
The first bound is obtained by taking the realization of the noise where $\br=\mathbf{0}$, in that case the input space partition is the entire space as any input is mapped to the same VQ code. As such, the mapping associated to this trivial partition has $\mathbf{0}$ slope (matrix filled with zeros) and a possibly nonzeros bias; as such the mapping is zero-dimensional (any points is mapped to the same point). This gives the lower bound stating that in the mixture of GDNs, one will have dimension $0$. For the other case, simply take the trivial case of $\br=\mathbf{1}$ which gives the result.
\end{proof}

\subsection{Proof of Lemma \ref{lemma:inverse}}
\label{proof:inverse}

\begin{proof}
First, as we impose injectivity, we can not have multiple regions of the input space, say $\omega$ and $\omega'$ such that $\bG(\omega) \cap \bG(\omega') \not = \emptyset$. Second, a region of the input space is mapped to another region in the output space by means of the affine transformation, thus even though the ambiant space $D$ might be of greater dimension that $\dim(\bG)$, the injectivity implies that points in $\omega$ are mapped to at most one point in  $\bG(\omega)$. They are affinely mapped meaning that the inverse is given by removing the bias and inverting the linear mapping which is given by the pseudo inverse. Recalling the above result on surjectivity, we see that for the GDN to be injective the per region dimension msut be $S$ showing existence of the pseudo inverse.
\end{proof}

\subsection{Proof of Proposition \ref{prop:disentenglement}}
\label{proof:disentenglement}

\begin{proof}
The proof is straightforward from the used definition of disentenglement. In fact, recall that we aim to have $\langle \bG(\bz)-\bG(\bz+\epsilon \delta_d),\bG(\bz)-\bG(\bz+\epsilon \delta_{d'})\rangle \approx 0$. In our case, consider only small transformation such that $\bz+\epsilon\delta_d$ and $\bz+\epsilon \delta_{d'}$ remain the in the same region $\omega$ in which was $\bz$. Then it is clear that for any positive constant $\epsilon$ fulfilling this condition, the above disentangelement definition translates into \begin{align*}
    \langle \bG(\bz)-\bG(\bz+\epsilon \delta_d),\bG(\bz)-\bG(\bz+\epsilon \delta_{d'})\rangle \approx 0 \iff \langle [\bA_{\omega}]_{.,d},[\bA_{\omega}]_{.,d'}\rangle \approx 0, 
\end{align*}
This gives a necessary condition which is not sufficient as this alone does not guarantee that each dimension of the latent space only impacts a single transformation of the output. But a disentangled representation must have near orthogonal columns for the slope matrices $\bA_{\omega}$.
\end{proof}

\subsection{Proof of Theorem \ref{thm:angle}}
\label{proof:angle}

\begin{proof}
First, notice that $P(\bA_{\omega}) = \bA_{\omega}(\bA_{\omega}^T\bA_{\omega})^{-1}\bA_{\omega}^T$ defines a projection matrix. In fact, we have that 
\begin{align*}
 P(\bA_{\omega})^2 &= \bA_{\omega}(\bA_{\omega}^T\bA_{\omega})^{-1}\bA_{\omega}^T\bA_{\omega}(\bA_{\omega}^T\bA_{\omega})^{-1}\bA_{\omega}^T\\
 &= \bA_{\omega}(\bA_{\omega}^T\bA_{\omega})^{-1}\bA_{\omega}^T\\
 &=P(\bA_{\omega})
\end{align*}
and we have that $(\bA_{\omega}^T\bA_{\omega})^{-1}$ is well defined as we assume injectivity ($\rank(\bA_{\omega})=S$) making the $S \times S$ matrix $\bA_{\omega}^T\bA_{\omega}$ full rank.
Now it is clear that this projection matrix maps an arbitrary point $\bx \in \mathbb{R}^D$ to the affine subspace $\bG(\omega)$ up to the bias shift. As we are interested in the angle between two adjacent subspaces $\bG(\omega)$ and $\bG(\omega')$ it is also clear that the biases (which do not change the angle) can be omited. Hence the task simplifies to finding the angle between $P(\bA_{\omega})$ and $P(\bA_{\omega'})$. This can be done by means of the greatest principal angle (proof can be found in \citet{stewart1973error}) with the result being
$\sin\big(\theta(\bG(\omega), \bG(\omega'))\big)= \| P(\bA_{\omega})- P(\bA_{\omega'})\|_2$ as desired.
\end{proof}

\subsection{Proof of Lemma~\ref{lemma:volume}}
\label{proof:volume}

\begin{proof}
In the special case of an affine transform of the coordinate given by the matrix $A \in \mathbb{R}^{D \times D}$ the well known result from demonstrates that the change of volume is given by
$|\det(A)|$ (see Theorem 7.26 in \cite{rudin2006real}). However in our case the mapping is a rectangular matrix as we span an affine subspace in the ambiant space $\mathbb{R}^D$ making $|\det(A)|$ not defined. However by applying Sard's theorem \cite{spivak2018calculus} we obtain that the change of volume from the region $\omega$ to the affine subspace $\bG(\omega)$ is given by $\sqrt{\det(A^TA)}$ which can also be written as follows with $USV^T$ the svd-decomposition of the matrix $A$:
\begin{align*}
    \sqrt{\det(A^TA)}=\sqrt{\det((USV^T)^T(USV^T))}
    =&\sqrt{\det((VS^TU^T)(USV^T))}\\
    =&\sqrt{\det(VS^TSV^T)}\\
    =&\sqrt{\det(S^TS)}\\
    =&\prod_{i:\sigma_i \not = 0}\sigma_i(A)
\end{align*}
\end{proof}

\subsection{Proof of Theorem~\ref{thm:general_density}}
\label{proof:general_density}

\begin{proof}
We will be doing the change of variables $\bz=(\bA^T_{\omega}\bA_{\omega})^{-1}\bA_{\omega}^T(\bx - \bb_{\omega})=\bA_{\omega}^+(\bx - \bb_{\omega})$, also notice that  $J_{\bG^{-1}}(\bx)=A^+$.
First, we know that
$
P_{\bG(\bz)}(\bx \in w)= P_{\bz}(\bz \in \bG^{-1}(w))= \int_{\bG^{-1}(w)}p_{\bz}(\bz) d\bz
$ which is well defined based on our full rank assumptions. We then proceed by
\begin{align*}
    P_{\bG}(\bx \in w)=& \sum_{\omega \in \Omega}\int_{\omega \cap w} p_{\bz}(\bG^{-1}(\bx)) \sqrt{ \det(J_{\bG^{-1}}(\bx)^TJ_{\bG^{-1}}(\bx))} d\bx\\
    =& \sum_{\omega \in \Omega}\int_{\omega \cap w}p_{\bz}(\bG^{-1}(\bx)) \sqrt{ \det((\bA_{\omega}^{+})^T\bA_{\omega}^{+})} d\bx\\
    =& \sum_{\omega \in \Omega}\int_{\omega \cap w}p_{\bz}(\bG^{-1}(\bx))  (\prod_{i:\sigma_i(\bA_{\omega}^+)>0}\sigma_i(\bA_{\omega}^+)) d\bx\\
    =& \sum_{\omega \in \Omega}\int_{\omega \cap w}p_{\bz}(\bG^{-1}(\bx)) (\prod_{i:\sigma_i(\bA_{\omega})>0}\sigma_i(\bA_{\omega}))^{-1} d\bx\;\;\;\text{Etape 1}\\
    =& \sum_{\omega \in \Omega}\int_{\omega \cap w} p_{\bz}(\bG^{-1}(\bx))\frac{1}{\sqrt{\det(\bA_{\omega}^T\bA_{\omega})}} d\bx\\
\end{align*}

Let now prove the Etape 1 step by proving that  $\sigma_i(A^{+})=(\sigma_i(A))^{-1}$ where we lighten notations as $A:=\bA_{\omega}$ and $USV^T$ is the svd-decomposition of $A$:
\begin{align*}
    A^{+}=(A^TA)^{-1}A^T
    =&((USV^T)^T(USV^T))^{-1}(USV^T)^T\\
    =&(VS^TU^TUSV^T)^{-1}(USV^T)^T\\
    =&(VS^TSV^T)^{-1}VS^TU^T\\
    =&V(S^TS)^{-1}S^TU^T\\
    \implies & \sigma_i(A^{+})=(\sigma_i(A))^{-1}
\end{align*}
with the above it is direct to see that $\sqrt{ \det((\bA_{\omega}^{+})^T\bA_{\omega}^{+})}=\frac{1}{\sqrt{\det(\bA_{\omega}^T\bA_{\omega})}}$ as follows
\begin{align*}
     \sqrt{ \det((\bA_{\omega}^{+})^T\bA_{\omega}^{+})}=\prod_{i:\sigma_i \not = 0}\sigma_i(\bA_{\omega}^{+})
     =&\prod_{i:\sigma_i \not = 0}\sigma_i(\bA_{\omega})^{-1}\\
     =&\left(\prod_{i:\sigma_i \not = 0}\sigma_i(\bA_{\omega})\right)^{-1}\\
     =& \frac{1}{\sqrt{\det(\bA_{\omega}^T\bA_{\omega})}}
\end{align*}
which gives the desired result.
\end{proof}

\subsection{Proof of Cor.~\ref{cor:entropy}}
\label{proof:entropy}
\begin{proof}
The derivation of the Entropy will consist in rewritting the Entropy w.r.t the distribution in the output space of the GDN and performing the change of coordinates leveragign the abvoe result to finally obtain the desired result as follows:
\begin{align*}
    E(\bp_{\bG}) =&-\sum_{\omega \in \Omega}\int_{\bG(\omega)}  \bp_{\bG}(\bx) \log (\bp_{\bG}(\bx)) d\bx \\
    =&-\sum_{\omega \in \Omega}\int_{\bG(\omega)} p_{\bz}(\bG^{-1}(\bx))\det(\bA_{\omega}^T\bA_{\omega})^{-\frac{1}{2}}  \log \left( p_{\bz}(\bG^{-1}(\bx))\det(\bA_{\omega}^T\bA_{\omega})^{-\frac{1}{2}} \right)\\
    =&-\sum_{\omega \in \Omega}\int_{\bG(\omega)} p_{\bz}(\bG^{-1}(\bx))\det(\bA_{\omega}^T\bA_{\omega})^{-\frac{1}{2}}  \left(\log \left( p_{\bz}(\bG^{-1}(\bx)) \right)+\log \left(\det(\bA_{\omega}^T\bA_{\omega})^{-\frac{1}{2}}\right) \right)\\
    =&-\sum_{\omega \in \Omega}\int_{\bG(\omega)} \det(\bA_{\omega}^T\bA_{\omega})^{-\frac{1}{2}} p_{\bz}(\bG^{-1}(\bx))  \log \left( p_{\bz}(\bG^{-1}(\bx)) \right)\\
    &-\sum_{\omega \in \Omega}\int_{\bG(\omega)} \det(\bA_{\omega}^T\bA_{\omega})^{-\frac{1}{2}} p_{\bz}(\bG^{-1}(\bx))  \log \left(
    \det(\bA_{\omega}^T\bA_{\omega})^{-\frac{1}{2}}\right) \\
    =&E(\bp_{\bz})\;\;\;\;\;\;\;\text{(apply the change of coordinate $\bz=G(\bx)$)}\\
    &-\sum_{\omega \in \Omega}\int_{\bG(\omega)} p_{\bz}(\bG^{-1}(\bx))\det(\bA_{\omega}^T\bA_{\omega})^{-\frac{1}{2}}  \log \left(\det(\bA_{\omega}^T\bA_{\omega})^{-\frac{1}{2}}\right)\\
    =&E(\bp_{\bz})+\frac{1}{2}\sum_{\omega \in \Omega}P(\bz \in \omega) \log \left(\det(\bA_{\omega}^T\bA_{\omega})\right)  \;\;\;\;\;\;\;\text{(apply the change of coordinate $\bz=G(\bx)$)}  
\end{align*}
which gives the desired result. For a complete review of integrals on manifold please see  \cite{cover2012elements}.
\end{proof}
\subsection{Proof of Theorem~\ref{thm:error_increase}}
\label{proof:error_increase}

\begin{proof}
For very small $N$ it is clear than in general, even if $S<S^*$, the memorization capacity of the generator will be s.t. it can fit through those points. Just imagine a couple of points sampled from a $2D$ linear manifold, even though $S=1$, the GDN can go through those two points and thus have $E^*=0$ for $N=2$. We now consider the case where $N$ is large enough. Two cases have to be studied.
\begin{itemize}
    \item Case $S<S^*$:  if $S<S^*$, the generated manifold can never be dense in the true linear manifold. This means that the newly introduced point will almost surely not lie in the span of the current generated manifold. Thus, $E^*(N+1)>E^*(N)$.
    
    \item Case $S\geq S^*$: in that case, it is clear the there always exist a setting of the parameters $\Theta$ s.t. the DGN spans the linear manifold. For example if using ReLU, consider any weights for the first $L-1$ layers s.t. the ReLU ae always ``on'' and use the last layer affine mapping to rotate and translate the affine subspace to the true one. That is, $E^*(N)=0, \forall N>0$.
\end{itemize}
The above demonstrates how for the simplest target manifold (linear) an too narrow DN leading to $S<S^*$ will haev a training error $E^*$ increasing with $N$ or $0$ if $S\geq S^*$ for any $N$.
\end{proof}

\subsection{Proof of Corollary~\ref{cor:gaussian_density}}
\label{proof:gaussian_density}
\begin{proof}

First, by applying the above results on the general density formula and setting $\bp_{\bz}$ a standard Normal distribution we obtain that
\begin{align*}
    \bp_{\bG}(\bx\in w)=&\sum_{\omega \in \Omega}\int_{\omega \cap w}\Indic_{\bx \in \bG(\omega)}p_{\bz}(\bG^{-1}(\bx)) \det(\bA_{\omega}^T\bA_{\omega})^{-\frac{1}{2}} d\bx\\
    =&\sum_{\omega \in \Omega}\int_{\omega \cap w}\Indic_{\bx \in \bG(\omega)}\frac{1}{(2\pi)^{S/2}\sqrt{\det(\bA_{\omega}^T\bA_{\omega})}}e^{-\frac{1}{2}\|\bG^{-1}(\bx) \|_2^2} d\bx\\
    =&\sum_{\omega \in \Omega}\int_{\omega \cap w}\Indic_{\bx \in \bG(\omega)}\frac{1}{(2\pi)^{S/2}\sqrt{\det(\bA_{\omega}^T\bA_{\omega})}}e^{-\frac{1}{2}((\bA^+_{\omega}(\bx-\bb_{\omega}))^T((\bA^+(\bx-\bb_{\omega}))} d\bx\\
    =&\sum_{\omega \in \Omega}\int_{\omega \cap w}\Indic_{\bx \in \bG(\omega)}\frac{1}{(2\pi)^{S/2}\sqrt{\det(\bA_{\omega}^T\bA_{\omega})}}e^{-\frac{1}{2}(\bx-\bb_{\omega})^T(\bA^+_{\omega})^T\bA^+_{\omega}(\bx-\bb_{\omega})} d\bx
\end{align*}
giving the desired result.
\end{proof}

\section{Codes of neighbour regions}
\label{appendix:neighbour_codes}

Each code is equivalent to a system of inequalities that define the regions. In fact, a code depends on the signs of the feature map pre activation (recall (\ref{eq:layer_mapping})). This defines a polytope in the input space and also in the output space. Now, when traveling from a point $\bz$ to another point $\bz'$ of a neighbouring region (recall Def.~\ref{def:adjacent}), we ask the question on how many indices of the code will change. That is, what is the Hamming distance between $\bq(\bz)$ and $\bq(\bz')$. As a neighbouring region is defined as a region which shares some of its boundary with another (their interior is disjoint) we can see that the degree of the face that is shared between the two regions define the amount of changes in their corresponding codes. If two regions share a $S-1$ dimensional face, then only $1$ value of the code changes. If they share in general a $S-r$ dimensional face, then the code will change by $r$ values. As most adjacent regions will share a high dimensional face, we see that $r$ tends to be small and thus codes are similar. For details and analytical study of the above please see \cite{lautensack_zuyev_2008}.

\section{More on Disentangled Latent Representations}
\label{sec:disentangled}

It has been coined that providing interpretable and practical generators lies in the ability to learn a {\em disentangled representation} of an input $\bx=\bG(\bz)$ \cite{schmidhuber1992learning,bengio2013representation}. The code $\bz$ should contain all the information present in $\bx$ in a compact
and interpretable structure where distinct,
informative factors of variations are encoded by different dimensions. Such motivations orginitated from the (non-)linear independent component analysis focusing on recovering independent factors from observed data \cite{comon1994independent,hyvarinen2016unsupervised}.
In fact, even in recent GAN/VAE based models, disentengled representations are associated to independent transformations of the input such as pose, hair color, eye color and so on which should behave independently form each other \cite{yim2015rotating,tran2017disentangled}. For a more in-depth review of learning disentangled representation, see \cite{locatello2018challenging}.

\section{Details on Training Procedure}
\label{appendix:error_with_n}

The experiment aims at depicting the training error being $E^*$ on the training set for varying latent dimensions $S$ in the simple case of a linear true data manifold approximation. In order to prevent any optimization unlucky degeneracy we repeat the training procedure $30$ times and compute for each poch the error $E^*$ and report the minimum over the $30$ trials and training epochs. We also set a very large number of epochs: $2000$. Due to the large number of trials and epochs the reported results are not due to some random initialization settings and convey the point of the result which is that even for such a simple data model (linear manifold) if $S<S^*$ then the training error $E^*$ will increase with $N$. Finally, the minimization over $z$ is replacer by an autoencoder with a very wide encoder s.t. it has the capacity for each training point to memorize the optimum $z$ that minimizes $E$. That is, when minimizing
\begin{align*}
    \min_{\Theta}\min_{\Theta'}\| \bG_{\Theta}(E_{\Theta'}(\bx))-\bx\| \approx \min_{\Theta}\min_{\bz}\| \bG_{\Theta}(\bz)-\bx\|,
\end{align*}
got a large enough encoder network $E$. In our case given that we used $S^*=6$ we used an encoder with $D_{\ell}=256$ units and $L=3$.

\section{More on Effect of Dropout/Dropconnect}
\label{sec:dropconnect}

As opposed to the Dropout case which applies the binary noise ont the feature maps $\bv_{\ell}$, Dropconnect \cite{wan2013regularization} applies this binary noise onto the slope matrices $\bW_{\ell}$ making the mapping noise become
\begin{align*}
    \bG(\bz) = \left(\prod_{\ell=L}^1\diag(\bq_{\ell})( \bW_{\ell}\odot\bR_{\ell})\right)\bz 
    + \sum_{\ell=1}^L  \left(\prod_{\ell'=L}^{\ell+1}\diag(\bq_{\ell'})(\bW_{\ell'}\odot\bR_{\ell'})\right)\bb_{\ell},
\end{align*}
where the binary noise matrices are denoted by $\bR_{\ell}$.
Despite this change of nosei application, the exact same result applies and Prop.~\ref{prop:mixture} also holds. That is the dropconnect equipped GDN becomes a mixture of GDNs with varying dimensions and parameters. Notice however that dropconnect will be less likely to reduce the ablated generator dimension as opposed to dropout due to its application on each entry of the weight matrix as opposed to an entire row at a time as depicted in Fig.~\ref{fig:dropout}.

\section{Acknowledgments }
This work was supported by 
NSF grants CCF-1911094, IIS-1838177, and IIS-1730574; 
ONR grants N00014-18-12571 and N00014-17-1-2551;
AFOSR grant FA9550-18-1-0478; 
DARPA grant G001534-7500; and a 
Vannevar Bush Faculty Fellowship, ONR grant N00014-18-1-2047, the Ken Kennedy Institute 2019/20 BP Graduate Fellowship.

\bibliography{references}
\bibliographystyle{icml2020}

\end{document}